\documentclass[pre,twocolumn,twoside,byrevtex,superscriptaddress,floatfix]{revtex4-1}

\usepackage{settings}

\setboolean{twocolswitch}{true}
\begin{document}

\title{\protect 


Interpretable bias mitigation for textual data: Reducing gender bias\\ in patient notes while maintaining classification performance 

















}
\author{
\firstname{Joshua R.}
\surname{Minot}
}
\email{joshua.minot@uvm.edu}
\affiliation{
  Computational Story Lab,
  The University of Vermont,
  Burlington, VT 05401.
} 
\affiliation{
  Vermont Complex Systems Center,
  The University of Vermont,
  Burlington, VT 05401.
} 
\affiliation{
  Department of Computer Science,
  The University of Vermont,
  Burlington, VT 05401.
}

\author{
\firstname{Nicholas}
\surname{Cheney}
}

\affiliation{
  Vermont Complex Systems Center,
  The University of Vermont,
  Burlington, VT 05401.
} 
\affiliation{
  Department of Computer Science,
  The University of Vermont,
  Burlington, VT 05401.
}
\affiliation{
  Neurobotics Lab,
  The University of Vermont,
  Burlington, VT 05401.
}


\author{
\firstname{Marc}
\surname{Maier}
}
\affiliation{
  MassMutual Data Science,
  Amherst, MA 01002.
} 

\author{
\firstname{Danne C.}
\surname{Elbers}
}
\affiliation{
  Computational Story Lab,
  The University of Vermont,
  Burlington, VT 05401.
} 
\affiliation{
  Vermont Complex Systems Center,
  The University of Vermont,
  Burlington, VT 05401.
} 
\affiliation{
  VA Cooperative Studies Program, VA Boston Healthcare System, 
  Boston, MA 02130.
} 

\author{
  \firstname{Christopher M.}
  \surname{Danforth}
}
\email{chris.danforth@uvm.edu}
\affiliation{
  Computational Story Lab,
  The University of Vermont,
  Burlington, VT 05401.
} 
\affiliation{
  Vermont Complex Systems Center,
  The University of Vermont,
  Burlington, VT 05401.
} 
\affiliation{
  Department of Computer Science,
  The University of Vermont,
  Burlington, VT 05401.
}

\affiliation{
  Department of Mathematics and Statistics,
  The University of Vermont,
  Burlington, VT 05401.
}

\author{
  \firstname{Peter Sheridan}
  \surname{Dodds}
}
\affiliation{
  Computational Story Lab,
  The University of Vermont,
  Burlington, VT 05401.
} 
\affiliation{
  Vermont Complex Systems Center,
  The University of Vermont,
  Burlington, VT 05401.
} 
\affiliation{
  Department of Computer Science,
  The University of Vermont,
  Burlington, VT 05401.
}
\affiliation{
  Department of Mathematics and Statistics,
  The University of Vermont,
  Burlington, VT 05401.
}

\date{\today}

\begin{abstract}
  \protect
  Medical systems in general, and patient treatment decisions and outcomes in particular, are affected by bias based on gender and other demographic elements. 
As language models are increasingly applied to medicine, there is a growing interest in building algorithmic fairness into processes impacting patient care. 
Much of the work addressing this question has focused on biases encoded in language models---statistical estimates of the relationships between concepts derived from distant reading of corpora. 
Building on this work, we investigate how word choices made by healthcare practitioners and language models interact with regards to bias. 
We identify and remove gendered language from two clinical-note datasets and describe a new debiasing procedure using BERT-based gender classifiers.
We show minimal degradation in health condition classification tasks for low- to medium-levels of bias removal via data augmentation. 
Finally, we compare the bias semantically encoded in the language models with the bias empirically observed in health records.
This work outlines an interpretable approach for using data augmentation to identify and reduce the potential for bias in natural language processing pipelines. 
\end{abstract}



\maketitle


\section{Introduction}\label{sec:introduction} 

Efficiently and accurately encoding patient information into medical records is a critical activity in healthcare. 
Electronic health records (EHRs) document symptoms, treatments, and other relevant histories---providing a consistent reference through disease progression, provider churn, and the passage of time.
Free-form text fields, the unstructured natural language components of a health record, can be incredibly rich sources of patient information. 
With the proliferation of EHRs, these text fields have also been an increasingly valuable source of data for researchers conducting large-scale observational studies. 

The promise of EHR data does not come without apprehension however, as the process of generating and analyzing text data is rife with potential for the influence of conscious and unconscious human bias. 
For example, health care providers entering information may have implicit or explicit demographic biases that ultimately become encoded in EHRs. 
Furthermore, language models that are often used to analyze clinical texts can encode broader societal biases~\cite{zhang2020hurtful}.
As patient data and advanced language models increasingly come into contact, it is important to understand how existing biases may be perpetuated in modern day healthcare algorithms.

In the healthcare context, many types of bias are worth considering. 
Race, gender, and socioeconomic status, among other attributes, all have the potential to introduce bias into the study and treatment of medical conditions. 
Bias may manifest in how patients are viewed, treated, and---most relevant here---documented. 
Due to ethical and legal considerations, as well as pragmatic constraints on data availability, we have focused the current research on gender bias.


There are many sources of algorithmic bias along with multiple definitions of fairness in machine learning~\cite{mehrabi2019survey}.
Bias in the data used for training algorithms can stem from imbalances in target classes, how specific features are measured, and historical forces leading certain classes to have longstanding, societal misrepresentation. 
Definitions of fairness include demographic parity, counterfactual fairness~\cite{kusner2017counterfactual}, and fairness through unawareness (FTU)~\cite{grgic2016case}. 
With our methods, we seek to provide human-interpretable insights on bias in the case of binary class data.
Further, using the same measurement, we experiment with the application of a FTU-like data augmentation process (although the concept of FTU does not neatly translate to unstructured text data).


There is growing interest in interpretable machine learning (IML) \cite{molnar2020interpretable}. 
In the context of deep language models this can involve interrogating the functionality of specific layers (e.g., BERTology \cite{rogers2020primer}), or investigating the impact of perturbations in data on outputs. This latter approach ties into the work outlined in this manuscript. 

Our use of the term `interpretable' here mostly refers to a more general case where a given result can be interpreted by a human reviewer. 
For instance, our divergence-based measures highlight gendered terms in plain English with a clearly explained ranking methodology. 
While this conceptualization is complementary to IML, it does not necessarily fit cleanly within the field---we will mention explicitly when referring to an IML concept.

\subsection{Prior work}

Gender bias in the field of medicine is a topic that must be viewed with nuance in light of the strong interaction between biological sex and health conditions. 
Medicine and gender bias interact in many ways---some of which are expected and desirable whereas others may have uncertain or negative impacts in patient outcomes.


Research has reported differences in the care and outcomes received by male and female patients for the same conditions. For example,
given the same severity symptoms, men have higher treatment rates for conditions such as coronary artery disease, irritable bowel syndrome, and neck pain~\cite{hamberg2008gender}. 
Women have higher treatment-adjusted excess mortality than men when receiving care for heart attacks \cite{alabas2017sex}.
Female patients treated by male physicians have higher mortality rates than when treated by female physicians---while male patients have similar mortality regardless of provider gender~\cite{greenwood2018patient}.

The rate of care-seeking behavior in men has been shown to be lower than women and has the potential to significantly affect health outcomes~\cite{galdas2005men}. 
Some work has shown female providers have higher confidence in the truthfulness of female patients and resulting diagnoses when compared to male providers \cite{gross2008association}.
The concordance of patient and provider gender is also positively associated with rates of cancer screening~\cite{malhotra2017impact}.

Beyond gender, the mortality rate of black infants has been found to be lower when cared for by black physicians rather than their white counterparts~\cite{greenwood2020physician}.
Race and care-seeking behavior have also been shown to interact, with black patients more often seeking cardiovascular care from black providers than non-black providers~\cite{alsan2018does}.  
It is important to note historical mistreatment and inequitable access when discussing racial disparities in health outcomes---for instance, the unethical Tuskegee Syphilis Study was found to lead to a 1.5-year decline in black male life expectancy through increased mistrust in the medical field after the exploitation of its participants was made public~\cite{alsan2018tuskegee}.


The gender of the healthcare practitioner can also impact EHR note characteristics that are subsequently quantified through language analysis tools.
The writings of male and female medical students have been shown to have differences, with female students expressing more emotion and male students using less space~\cite{lin2016word}. 
More generally, some work has shown syntactic parsers generalize well for men and women when trained on data generated by women whereas training the tools on data from men leads to poor performance for texts written by women~\cite{garimella2019women}. 


The ubiquity of text data along with advances in natural language processing (NLP) have led to a proliferation of text analysis in the medical realm. 
Researchers have used social media platforms for epidemiological research~\cite{prieto2014twitter, rodriguez2018twitter, salathe2018digital}---raising a separate set of ethical concerns~\cite{mello2020ethics}.
NLP tools have been used to generate hypotheses for biomedical research~\cite{sybrandt2017moliere}, 
detect adverse drug reactions from social media~\cite{yang2012social}, 
and expand the known lexicon around medical topics~\cite{percha2018expanding, fan2019using}. 
There are numerous applications of text analysis in medicine beyond patient health records. 
While this manuscript does not directly address tasks outside of clinical notes, it is our hope that the research could be applied to other areas.
It is because our methods are interpretable and based on gaining an empirical view of bias that we feel they could be a first resource in understanding bias beyond our example cases of gender in clinical texts.


Our work leverages computational representations of statistically derived relationships between concepts, commonly  known as \textit{word embedding} models \cite{morin2005hierarchical}.
These real-valued vector representations of words facilitate comparative analyses of text data with machine learning methods. 
The generation of these vectors depends on the distributional hypothesis, which states that similar words are more likely to appear together within a given context. 
Ideally, word embeddings map semantically similar words to similar regions in the vector space---or `semantic space' in this case.
The choice of training dataset heavily impacts the qualities of the language model and resulting word embeddings. 
For instance, general purpose language models are often trained on Wikipedia and the Common Crawl collection of web pages (e.g., BERT~\cite{devlin2018bert}, RoBERTa~\cite{liu2019roberta}). 
Training language models on text from specific domains often improves performance on tasks in those domains (see below). 
More recent, state-of-the-art word embeddings (e.g., ELMo~\cite{peters2018deep}, BERT~\cite{devlin2018bert}, \mbox{GPT-2}~\cite{radford2019language}) are generally `contextual', where the vector representation of a word from the trained model is dependent on the context around the word. 
Older word embeddings, such as GloVe~\cite{pennington2014glove} and word2vec~\cite{mikolov2013efficient, mikolov2013distributed}, are `static', where the output from the trained model is only dependent on the word of interest---with context still being central to the task of training the model.


As medical text data are made increasingly accessible through EHRs, there has been a growing focus on developing word embeddings tailored for the medical domain.
The practice of publicly releasing pre-trained, domain-specific word embeddings is common across domains, and it can be especially helpful in medical contexts described using specialized vocabulary (and even manner of writing). 
SciBERT is trained on a random sample of over one million biomedical and computer science papers~\cite{beltagy2019scibert}. 
BioBERT similarly is trained on papers from PubMed abstracts and articles~\cite{lee2020biobert}. 
There are also pre-trained embeddings focused on tasks involving clinical notes. 
Clinical BERT~\cite{alsentzer-etal-2019-publicly,huang2019clinicalbert} is trained on clinical notes from the \MIMIC{} dataset~\cite{johnson2016mimic}.
A similar approach was applied with the XLNet architecture, resulting in clinical XLNet~\cite{huang2019clinical}. 
These pre-trained embeddings perform better on domain-specific tasks related to the training data and procedure.

The undesirable bias present in word embeddings has attracted growing attention in recent years. 
Bolukbasi \textit{et al.} present evidence of gender bias in word2vec embeddings, along with proposing a method for removing bias from gender-neutral terms~\cite{bolukbasi2016man}.
Vig \textit{et al.} investigate which model components (attention heads) are responsible for gender bias in transformer-based language models (GPT-2)~\cite{vig2020investigating}.
A simple way to mitigate gender bias in word embeddings is to `swap' gendered terms in training data when generating word embeddings~\cite{zhao2018learning}. 
Beutel \textit{et al.} \cite{beutel2017data} develop an adversarial system for debiasing language models---in the process, relating the distribution of training data to its effects on properties of fairness in the adversarial system.
Simple masking of names and pronouns may reduce bias and improve classification performance for certain language classification tasks~\cite{dayanik2020masking}. 
Some of these techniques for bias detection and mitigation have been critiqued as merely capturing over-simplified dimensions of bias---with proper debiasing requiring more holistic evaluation~\cite{gonen2019lipstick}. 

Data augmentation has been used to improve classification performance and privacy of text data. 
Simple methods include random swapping of words, random deletion, and random insertion~\cite{wei2019eda}.
More computationally expensive methods may involve using language models to generate contextually accurate synonyms~\cite{kobayashi2018contextual}, or even running text through multiple rounds of machine translation (e.g., English text to French and back again)~\cite{yu2018qanet}.
De-identification is perhaps the most common data augmentation task for clinical text.
Methods may range from simple dictionary look-ups~\cite{meystre2010automatic} to more advanced neural network approaches~\cite{dernoncourt2017identification}. 
De-identification approaches may be too aggressive and limit the utility of the resulting data while also offering no formal privacy guarantee.
The field of differential privacy~\cite{dwork2008differential} offers principled methods for adding noise to data, and some recent work has explored applying these principles to text data augmentation~\cite{adelani2020privacy}.
Applying data-augmentation techniques to pipelines that use contextual word-embeddings presents some additional uncertainty given on-going nature of research working on establishing what these trained embeddings actually represent and how they use contextual clues (e.g., the impact of word order on downstream tasks~\cite{pham2020out}).

In the present study, we explore the intersection of the bias that stems from language choices made by healthcare providers and the bias encoded in word embeddings commonly used in the analysis of clinical text. 
We present interpretable methods for detecting and reducing bias present in text data with binary classes.
Part of this work is investigating how orthogonal text relating to gender bias is to text related to clinically-relevant information.
While we focus on gender bias in health records, this framework could be applied to other domains and other types of bias as well. 
In Sec.~\ref{sec:methods}, we describe our data and methods for evaluating bias. 
In Sec.~\ref{sec:results}, we present our results contrasting empirically observed bias in our sample data with bias encoded in word embeddings.
Finally, in Sec.~\ref{sec:conclusion}, we discus the implications of our work and potential avenues for future research.

\section{Methods}\label{sec:methods}
Here we outline our methods for identifying and removing gendered language and evaluating the impact of this data augmentation process.
We also provide brief descriptions of the datasets for our case study.
The bias evaluation techniques fall into two main categories. 
First, there are extrinsic evaluation tasks focused on comparing the performance of classifiers as we vary the level of data augmentation.
For our dataset, this process involves testing health-condition and gender classifiers on augmented data. 
The extrinsic evaluation is meant to be similar to some real-world tasks that may utilize similar classifiers. 
Second, we make intrinsic evaluations of the language by looking at bias within word-embedding spaces and empirical word-frequency distributions of the datasets. 
The set of methods presented here enable the identification of biased language usage, a data augmentation approach for removing this language, and an example benchmark for evaluating the performance impacts on two biomedical datasets.

\subsection{Rank-divergence and trimming}

We parse clinical notes into $n$-grams---sequences of space delimited strings such as words or sentence fragments---and generate corresponding frequency distributions.  
To quantify the bias of specific $n$-grams we compare their frequency of usage in text data corresponding to each of our two classes.
The same procedure is extended as a data augmentation technique intended to remove biased language in a targeted and principled manner.

\begin{figure*}[ht]
    \centering
    \includegraphics[width=\textwidth]{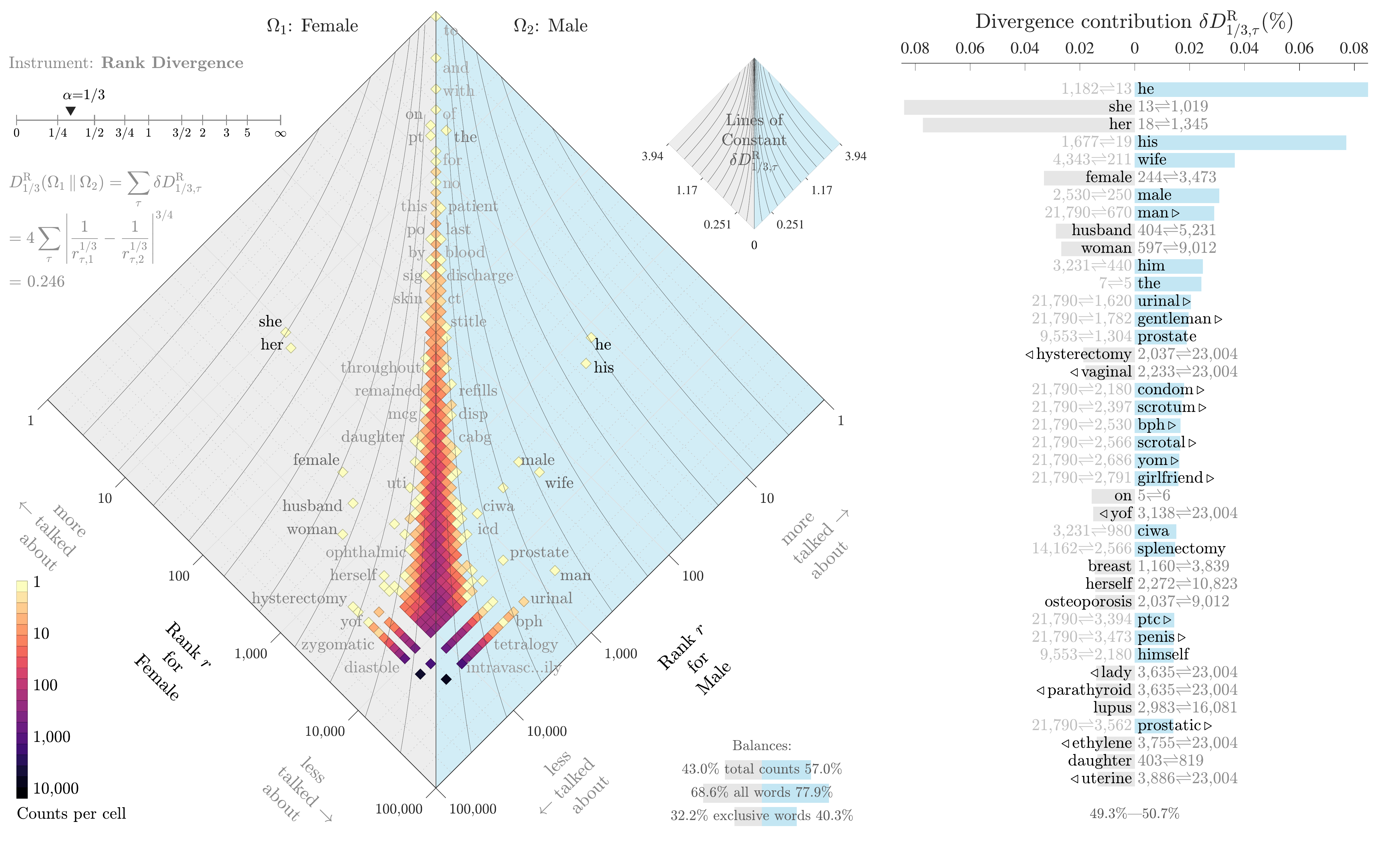}
    \caption{\textbf{Rank-turbulence divergence allotaxonograph} \cite{dodds2020allotaxonometry} \textbf{for male and female documents in the \MIMIC dataset.} 
    For this figure, we generated 1-gram frequency and rank distributions from documents corresponding to male and female patients.
    Pronouns such as ``she'' and ``he'' are immediately apparent as drivers of divergence between the two corpora.
    From there, the histogram on the right highlights gendered language that is both common and medical in nature. 
    Familial relations (e.g., ``husband'' and ``daughter'') often present as highly gendered according to our measure.
    Further, medical terms like ``hysterectomy'' and ``scrotum'' are also highly ranked in terms of their divergence.
    In the main horizontal bar chart, the numbers next to the terms represent their rank in each corpus, while terms that appear in only one corpus are indicated with a rotated triangle.   
    The smaller three vertical bars describe balances between the male and female corpora: 
    43\% of total 1gram counts appear in the female corpus;
    we observed 68.6\% of all 1grams in the female corpus; 
    and 32.2\% of the 1grams in the female corpus are unique to that corpus.
    }
    \label{fig:rtd_1gram_mimic}
\end{figure*}

More specifically, in our case study of gendered language in clinical notes we quantify the ``genderedness'' of $n$-grams by comparing their frequency of usage in notes for male and female patient populations. 
For the task of comparing frequency of $n$-gram usage we use rank-turbulence divergence (RTD), as defined by Dodds \textit{et al.}~\cite{dodds2020allotaxonometry}. 
The rank-turbulence divergence between two sets, $\Omega_1$ and $\Omega_2$, is calculated as follows,

\begin{equation}
\begin{aligned}
D^{R}_{\alpha}(\Omega_1 || \Omega_2) 
&= \sum \delta D^{R}_{\alpha,\tau}  \\
&= \frac{\alpha +1}{\alpha} \sum_\tau \left| \frac{1}{r_{\tau,1}^\alpha} - \frac{1}{r_{\tau,2}^\alpha} \right| ^{1/(\alpha+1)} \,,
\end{aligned}
\end{equation}

where $r_{\tau,s}$ is the rank of element $\tau$ ($n$-grams, in our case) in system $s$ and $\alpha$ is a tunable parameter that affects the starting and ending ranks. 
While other techniques could be used to compare the two $n$-gram frequency distributions, we found RTD to be robust to differences in overall volume of $n$-grams for each patient population. 
For example, Fig.~\ref{fig:rtd_1gram_mimic} shows the RTD between 1-grams from clinical notes corresponding to female and male patients.

A brief note on notation: $\tau$ always represents a unique element, or $n$-gram in our case. 
In certain contexts $\tau$ may be an integer value that ultimately maps back to the element's string representation.
This integer conversion is to allow for clean indexing---in these cases $\tau$ can be converted back to a string representation of the element with the array of element strings $W_\tau$. 

\begin{figure*}
    \centering
    \includegraphics[width=\textwidth]{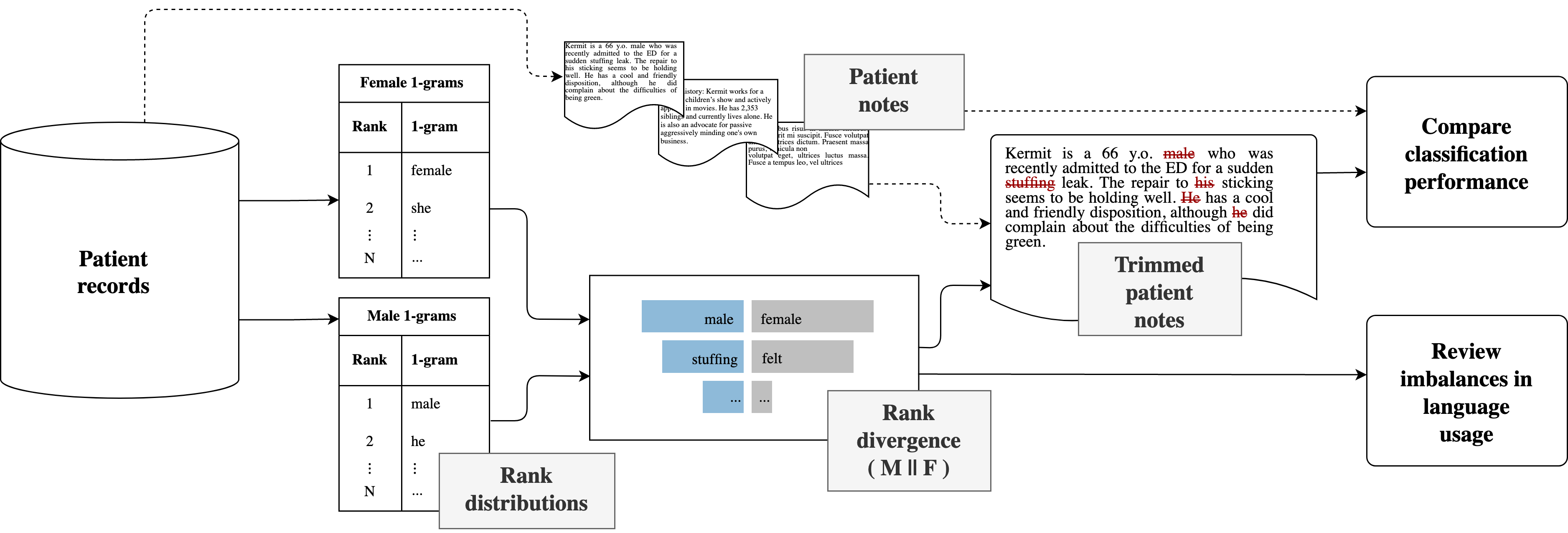}
    \caption{
    \textbf{Overview of the rank-turbulence divergence trimming procedure.}
    Solid lines indicate steps that are specific to our trimming procedure and evaluation process. 
    The pipeline starts with a repository of patient records that include clinical notes and class labels (in our case gender and ICD9 codes). 
    From these notes we generate $n$-gram rank distributions for the female and male patient populations, which are then used to calculate the rank-turbulence divergence (RTD) for individual $n$-grams. 
    Sorting the $n$-grams based on RTD contribution, we then trim the clinical notes. 
    Finally, we view the results directly from the RTD calculation to review imbalance in language use. 
    With the trimmed documents we compare the performance of classifiers on both the un-trimmed notes and notes with varying levels of trimming applied. 
    }
    \label{fig:trim_diagram}
\end{figure*}

We use the individual rank-turbulence divergence contribution, $\delta D^{R}_{\alpha,\tau}$, of each $1$-gram to the gendered divergence, $D^{R}_{\alpha}(\Omega_\textnormal{female} || \Omega_\textnormal{male})$, to select which terms to remove from the clinical notes. 
First, we sort the $1$-grams based on their rank-turbulence divergence contribution. 
Next, we calculate the cumulative proportion of the overall rank-turbulence divergence, $RC_\tau$, that is accounted for as we iterate through the sorted $1$-gram list from words with the highest contribution to the least contribution (in this case, terms like ``she'' and ``gentleman'' will tend to have a greater contribution). 
Finally, we set logarithmically spaced thresholds of cumulative rank-divergence values to select which $1$-grams to trim. 
The method allows us to select sets of $1$-grams that contribute the most to the rank-divergence values (measured as divergence per $1$-gram). Fig.~\ref{fig:trim_diagram} provides a graphical overview of this procedure. 

\begin{figure}[h]
    \centering
    \includegraphics[width=\columnwidth]{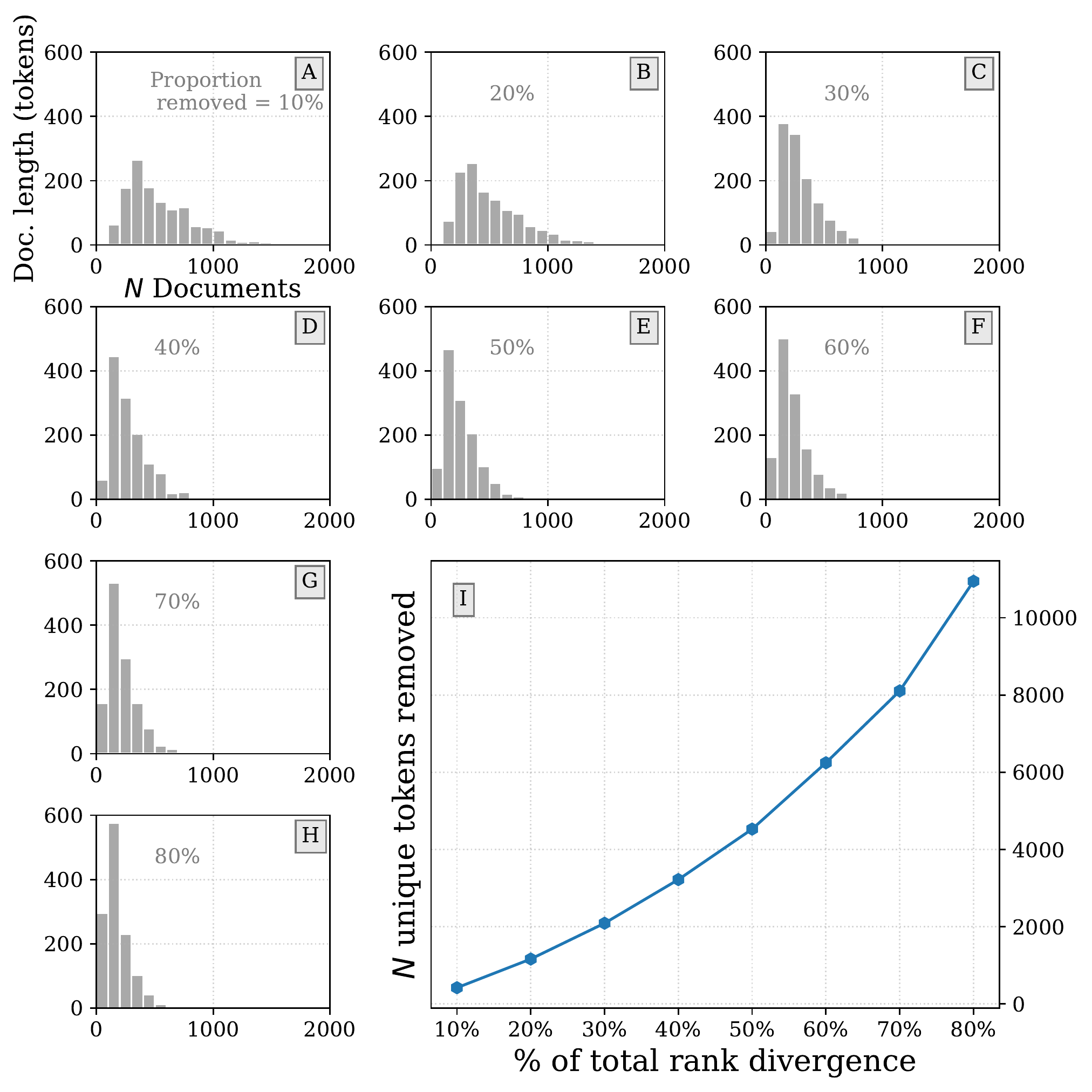}
    \caption{\textbf{Document length after applying a linearly-spaced rank-turbulence divergence based trimming procedure.}
    Percentage values represent the percentage of total rank-turbulence divergence removed.
    Trimming is conducted by sorting words highest-to-lowest based on their individual contribution to the rank-turbulence divergence between male and female corpora (i.e., the first 10\% trim will include words that, for most distributions, contribute far more to rank-turbulence divergence than the last 10\%).}
    \label{fig:doclen}
\end{figure}

Using this selection criteria, we are able to remove the least number of $1$-grams per a given amount of rank-turbulence divergence removed from the clinical notes. 
The number of unique $1$-grams removed per cumulative amount of rank-turbulence divergence grows super linearly as seen in Fig.~\ref{fig:doclen}I. 
This results in relatively stable distributions of document lengths for lower trim values (10--30\%), although at higher trim values the procedure drastically shrinks the size of many documents (Fig.~\ref{fig:doclen}A-H).

To implement this trimming procedure, we use regular expressions to replace the $1$-grams we have identified for removal with a space character. 
Our string removal procedure is applied to the overall corpus of data, upstream of any train-test dataset generation for specific classification tasks. 

Other potential string replacement strategies include redaction with a generic token or randomly swapping $n$-grams that appear within the same category across the corpus~\cite{adelani2020privacy}.
The RTD method we use could also be adapted for use with these and other replacement strategies. 
We chose string removal because of its simplicity and prioritization of the de-biasing task over preserving semantic structure (i.e., it presents an extreme case of data augmentation). 
The pipeline's performance on downstream tasks provides some indication of the semantic information retained, and as we show in Sec.~\ref{sec:results} it is possible retain meaningful signals while pursuing relatively aggressive string removal.

\begin{algorithm}[H] 
\caption{RTD trimming procedure}
\label{alg:rtd_trim}
\begin{algorithmic}[1]
    \Require{Documents $D_i \qquad ,i = 1,...,N  $}
    \Require{1-gram rank dists. for each class $\Omega_\psi \qquad, \psi = 1,2 $}
    \Ensure{Trimmed text data $C_i^{(k)} \qquad ,i = 1,...,N \,; k\in(0,1] $}
    \State{$\delta D^{R}_{\alpha,\tau},  W_\tau \gets \verb|RTD_calc|(\Omega_1, \Omega_2,\alpha)$ \qquad, $ \tau = 1,...,M$} 
    \Comment{ $\delta D^{R}_{\alpha,\tau}$ is RTD contribution for ngram $W_\tau$, both sorted by RTD contribution}
    \State{$RC_\tau \gets \verb|cumsum|(\delta D^{R}_{\alpha,\tau})$}
    \For{$k = .1,.2,...,.9 $}
    \State{$r \gets  \verb|max|(\verb|where|(RC<=b)) $}
    \Comment{index up to bin max $b$}
    \State{$S \gets W_{0:r}$}
    \For{$i = 1,...,N$}
    \State{$C_i^{(k)} \gets \verb|strip|(D_i, S)$}
    \Comment{remove $1-$grams from doc.}
    \EndFor
    \EndFor
\end{algorithmic} 
\end{algorithm}

\subsection{Language models}
\label{sec:language_models}

Large language models are increasingly common in many NLP tasks, and we feel it is important to present our results in the context of a pipeline that utilizes these models.
Furthermore, language models have the potential to encode bias, and we found it necessary to contrast our empirical bias detection methods with bias metrics calculated on general purpose and domain-adapted language models. 

We use pre-trained BERT-base~\cite{devlin2018bert} and Clinical BERT~\cite{alsentzer-etal-2019-publicly} word embeddings. 
BERT provides a contextual word embedding trained on ``general'' language whereas Clinical BERT builds on these embeddings by utilizing transfer learning to improve performance on scientific and clinical texts. 
All models were implemented in PyTorch using the Transformers library~\cite{Wolf2019HuggingFacesTS}. 

\begin{figure}[h]
    \centering
    \includegraphics[width=\columnwidth]{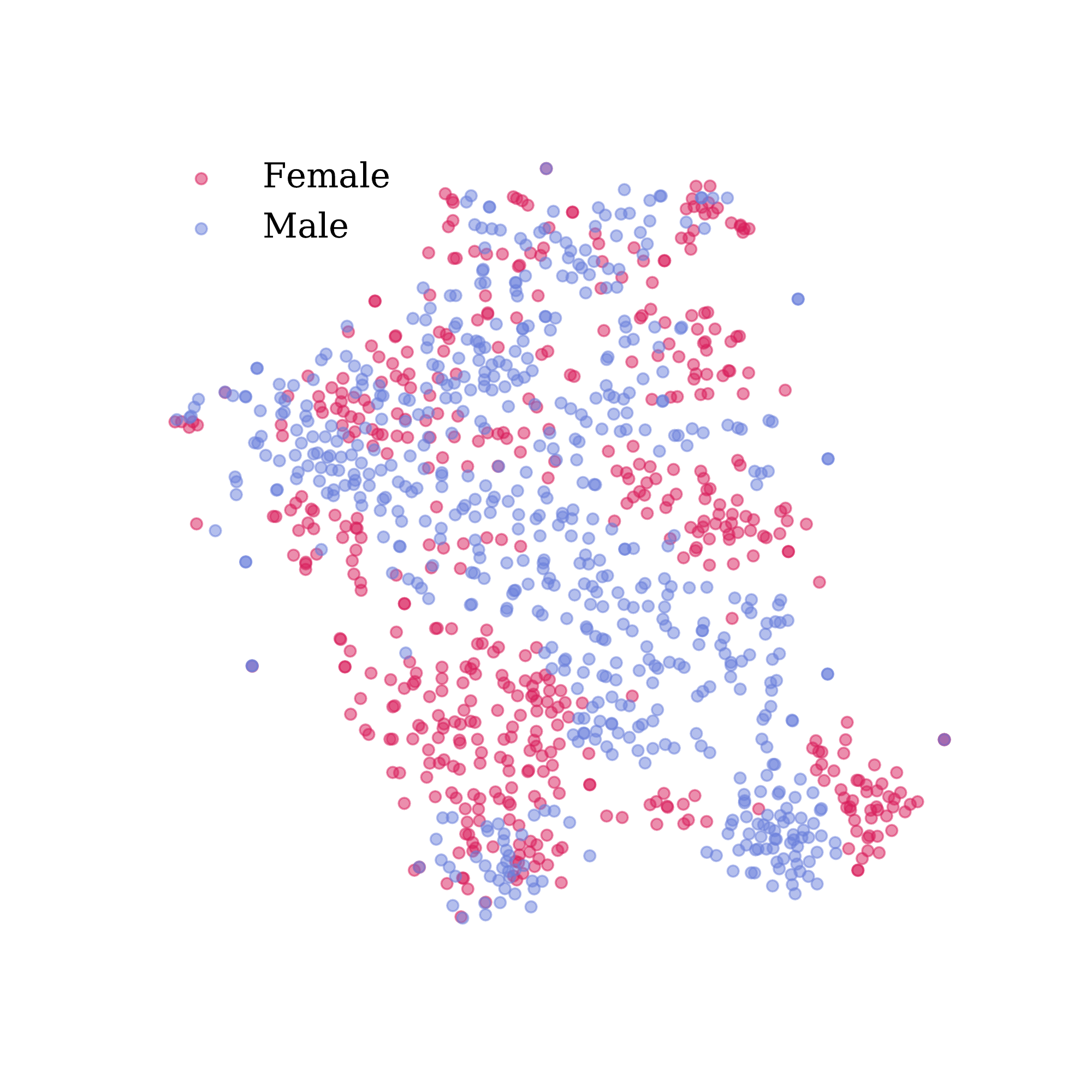}
    \caption{\textbf{A tSNE embedding of  \iibb  document vectors generated using a pre-trained version of BERT with off-the-shelf weights.}
    We observe the appearance of gendered clusters even before training for a gender classification task. 
    See Fig.~\ref{fig:tSNE_ClinicalBERT_i2b2} for the same visualization but with Clinical BERT embeddings.  }
    \label{fig:tSNE_BERT_i2b2}
\end{figure}

For tasks such as nearest-neighbor classification and gender-similarity scoring, we use the off-the-shelf weights for BERT and Clinical BERT (see Fig.~\ref{fig:tSNE_BERT_i2b2} for an example of \iibb embedding space). 
These models were then fine-tuned on the gender and health-condition classification tasks.

In cases where we fine-tuned the model, we added a final linear layer to the network. 
All classification tasks were binary with a categorical cross-entropy loss function. 
All models were run with a maximum sequence length of 512, batch size of 4, and gradient accumulation steps set to 12.
We considered various methods for handling documents longer than the maximum sequence length (see \textit{Variable length note embedding} in SI), but ultimately the performance gains did not merit further use.

We also run a nearest-neighbor classifier on the document embeddings produced by the off-the-shelf BERT-base and Clinical BERT models. 
This is intended to be a point of comparison when evaluating the performance of bias present within the embedding space, as indicated by performance on extrinsic classification tasks. 

In addition to the BERT-based language models, we used a simple term frequency-inverse document frequency (TFIDF) \cite{robertson2004understanding} based classification model as a point of comparison. 
For this model, we fit a TFIDF vectorizer to our training data and use logistic regression for binary classification. 

\subsection{Gender distances in word embeddings}

Using a pre-trained BERT model, we embed all the 1-grams present in the clinical note datasets. 
For this task, we retain the full length vector for each 1-gram, taking the average in cases where additional tokens are created by the tokenizer. 
The results of this process are 1x768 vectors for each $n$-gram. 
We also calculate the average embedding for a collection of terms manually selected to constitute `gender clusters'. 
From these gender clusters we calculate the cosine similarly to each of the embeddings for $n$-grams in the Zipf distribution. 

Using measures such as cosine similarity with BERT raises some concerns, especially when looking at absolute values. 
BERT was not trained on the task of calculating sentence or document similarities. 
With BERT-base, all dimensions are weighted equally, which when applying cosine similarity can result in somewhat arbitrary absolute values. 
As a workaround, we believe that using the ranked value of word and document embeddings can produce more meaningful results (if we do not wish to fine-tune BERT on the task of sentence similarity). 
We use both absolute values and ranks of cosine similarity when investigating bias in BERT-based language models---finding the absolute values of cosine similarity to be meaningful in our relatively coarse-grained analysis.
Further, taking the difference in cosine similarities for each gendered cluster addresses some of the drawbacks of examining cosine similarity values in pre-trained models. 

Generating word or phrase embeddings from contextual language models raises some challenges in terms of calculating accurate embedding values.
In many cases, the word-embedding for a given $1$-gram---produced by the final layer of a model such as BERT---can vary significantly depending on context~\cite{ethayarajh2019contextual}.
Some researchers have proposed converting contextual embeddings to static embeddings to address this challenge~\cite{bommasani2020interpreting}. 
Others have presented methods for creating template sentences and comparing the relative probability of masked tokens for target terms~\cite{kurita2019measuring}. 
After experimenting with the template approach, we determined that the resulting embeddings were not different enough to merit switching away from the simple isolated 1-gram embeddings. 

\subsection{Rank-turbulence divergence for embeddings and documents}

We use rank-turbulence divergence in order to compare bias encoded in word-embeddings and empirical data.
For word embeddings, we need to devise a metric for bias---here we use cosine similarity between biased-clusters and candidate $n$-grams.
The bias in the empirical data is evaluated using RTD for word-frequency distributions corresponding to two labeled classes. 

In terms of the clinical text data, for the word embeddings we use cosine similarity scores to evaluate bias relative to known gendered $n$-grams.
For the clinical note datasets (text from documents with gender labels), we use rank-turbulence divergence calculated between the male and female patient populations. 

To evaluate bias in the embedding space, we rely on similarity scores relative to known gendered language.
First, we create two gendered clusters of $1$-grams---these clusters represent words that are manually determined to have inherent connotations relating to female and male genders.
Next, we calculate the cosine similarity between the word embeddings for all $1$-grams appearing in the empirical data and the average vector for each of the two gendered clusters. 
Finally, we rank each $1$-gram based on the distribution of cosine similarity scores for the male and female clusters.

For the empirical data, we calculate the RTD for $1$-grams appearing in the clinical note data sets. 
The RTD value provides an indication of the bias---as indicated by differences in specific term frequency---present in the clinical notes.

Combined, these steps provide ranks for each $1$-gram in terms of how much it differentiates the male and female clinical notes.
Here again we can use the highly flexible rank-turbulence divergence measure to identify where there is `disagreement' between the ranks returned by evaluating the embedding space and ranks from the empirical distribution. 
This is a divergence-of-divergence measure, using the iterative application of rank-turbulence divergence to compare two different measures of rank. 
Going forward, we refer to this measure as RTD$^2$. 
RTD$^2$ provides an indication of which $n$-grams are likely to be reported as less gendered in either the embedding space or in the empirical evaluation of the documents. 
For our purposes, RTD$^2$ is especially useful for highlighting $n$-grams that embedding-based debiasing techniques may rank as minimally biased, despite the emperical distribution suggesting otherwise.

\subsection{Data}


We use two open source datasets for our experiments: the n2c2 (formerly i2b2) 2014 deidentification challenge~\cite{kumar2015creation} 
and the \MIMIC{} critical care database~\cite{johnson2016mimic}. The \iibb data comprises around 1300 documents with no gender or health condition coding (we generate our own labels for the former). 
\MIMIC{} is a collection of diagnoses, procedures, interventions, and doctors notes for 46,520 patients that passed through an intensive care unit. 
There are 26,121 males and 20,399 females in the dataset, with over 2 million individual documents. 
\MIMIC{} includes coding of health conditions with International Classification of Diseases (ICD-9) codes, as well as patient sex.  

\begin{figure}[h]
    \centering
    \includegraphics[width=\columnwidth]{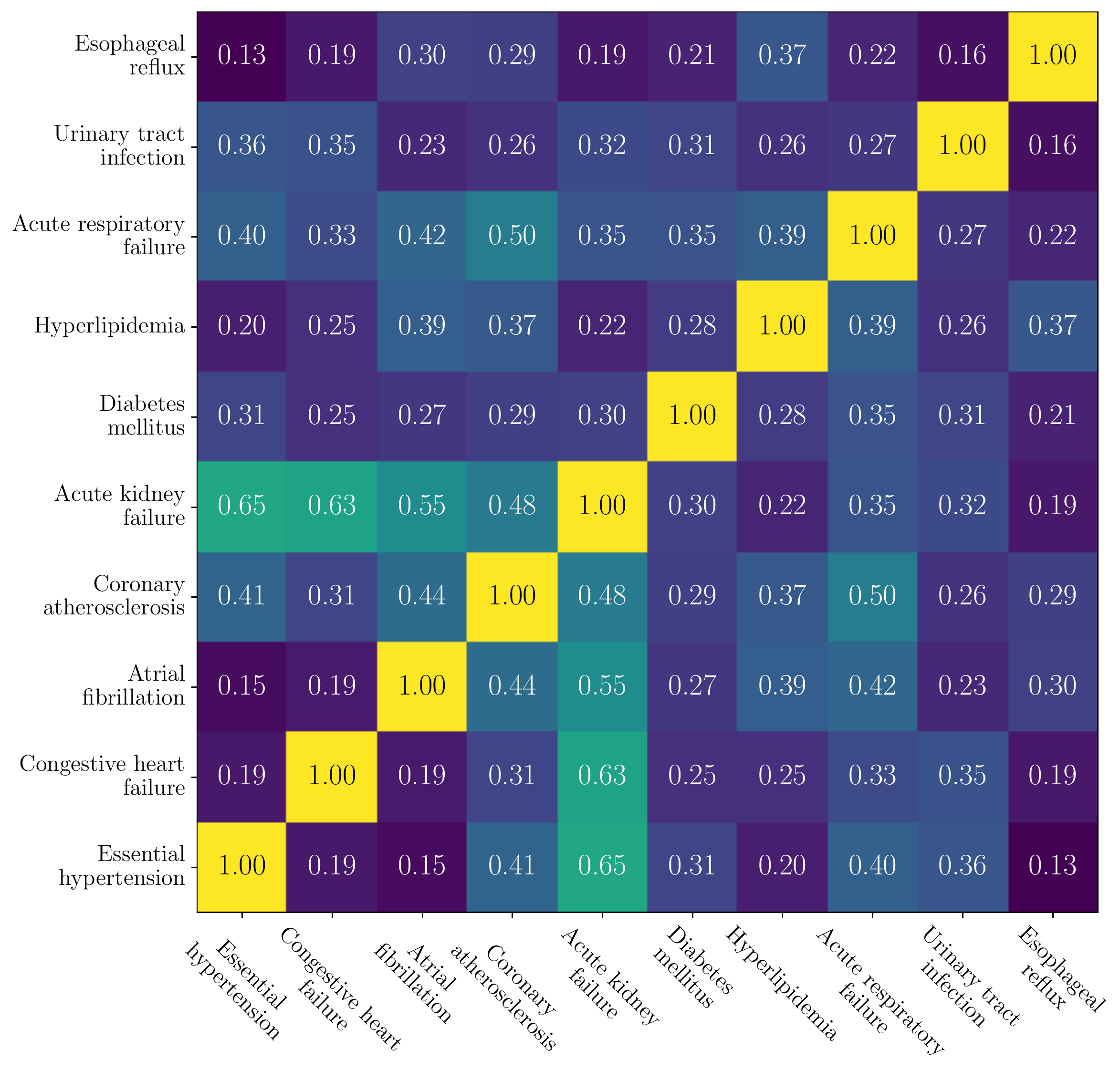}
    \caption{\textbf{Normalized rates of health-condition co-occurrence for the top 10 ICD-9 codes. } }
    \label{fig:cond_matrix}
\end{figure}

For \MIMIC, we focus our health-condition classification experiments on records corresponding to patients with at least 1 of the top 10 most prevalent ICD-9 codes. 
We restrict our sample population to those patients with at least one of the 10 most common health conditions---randomly drawing negative samples from this subset for each condition classification experiment. 
Rates of coincidence vary between 0.65 and 0.13 (Fig.~\ref{fig:cond_matrix}).
All but one of the top 10 health conditions have more male than female patients (Table~\ref{tab:mimic_sex}). 
As a point of reference, we also present summary results for records corresponding to patients with ICD-9 codes that appear at least 1000 times in the \MIMIC{} data (Table~\ref{tab:top1000MIMIC}).

\subsection{Text pre-processing}

Before analyzing or running the data through our models, we apply a simple pre-processing procedure to the text fields of the \iibb and \MIMIC{} data sets. 
We remove numerical values, ranges, and dates from the text. 
This is done in an effort to limit confounding factors related to specific values and gender (e.g., higher weights and male populations).
We also strip some characters and convert common abbreviations. 
See Sec.~\ref{sec:SI_note_selection} for information on note selection.

\begin{table}
\rowcolors{2}{lightgray}{white}
\begin{tabular}{l|r|r|r|r}
\toprule
{} & \multicolumn{2}{l}{\textbf{Sex Count}} & {} \\
 \textbf{ICD Description.}  &     \textbf{ $N_\textnormal{f}$} & \textbf{$N_\textnormal{m}$} & \textbf{ $N_\textnormal{f}$/$N_\textnormal{total}$} & \textbf{ $N_\textnormal{m}$/$N_\textnormal{total}$} \\ 
 \hline
\midrule
Acute kidney failure      &   3941 &   5178 &  0.43 &  0.57 \\
Acute respiratory failure &   3473 &   4024 &  0.46 &  0.54 \\
Atrial fibrillation       &   5512 &   7379 &  0.43 &  0.57 \\
Congestive heart failure  &   6106 &   7005 &  0.47 &  0.53 \\
Coronary atherosclerosis  &   4322 &   8107 &  0.35 &  0.65 \\
Diabetes mellitus         &   3902 &   5156 &  0.43 &  0.57 \\
Esophageal reflux         &   2990 &   3336 &  0.47 &  0.53 \\
Essential hypertension    &   9370 &  11333 &  0.45 &  0.55 \\
Hyperlipidemia            &   3537 &   5153 &  0.41 &  0.59 \\
Urinary tract infection   &   4027 &   2528 &  0.61 &  0.39 \\

\bottomrule
\end{tabular}
\caption{\textbf{Patient sex ratios for the top 10 conditions in \MIMIC.}
For most health conditions there is an a imbalance in the gender ratio between male and female patients.
This reflects an overall bias in the \MIMIC dataset with has more male patients.}
\label{tab:mimic_sex}
\end{table}

\section{Results and Discussion}\label{sec:results}
Here we present the results of applying empirical bias detection and mitigation methods. 
Using rank-turbulence divergence (RTD) we rank $n$-grams based on their contribution to bias between two classes. 
Next we apply a data augmentation procedure where we remove 1-grams based on their ranking in the RTD results. 
The impact of the data augmentation process is measured by tracking classification performance as we apply increasingly aggressive 1-gram trimming to our clinical note datasets.
Finally, we compare the bias present in BERT with the empirical bias we detect in the case study datasets. 

One of our classification tasks is predicting patient gender from EHR notes.
We include gender classification as a synthetic test that is meant to directly indicate the gender signal present in the data.
Gender classification is an unrealistic task that we would not expect to see in real-world applications, but serves as an extreme case that provides insight on the potential for other classifiers to incorporate gender information.

We present results for both the \iibb and \MIMIC{} datasets. 
The \iibb dataset provides a smaller dataset with more homogeneous documents and serves as a reference point for the tasks outlined here. 
\MIMIC{} is much larger and its explicit coding of health conditions allows us to bring the extrinsic task of condition classification into our evaluation of bias and data augmentation.

\subsection{Gender divergence}

Interpretability is a key facet of our approach to empirical bias detection.
To gain an understanding of biased language usage we start by presenting the ranks of RTD values for individual $n$-grams in text data corresponding to each of the binary classes. 
The allotaxonographs we use to present this information (e.g., Fig.~\ref{fig:rtd_1gram_mimic}) show both the RTD values and a 2-d rank-rank histogram for $n$-grams in each class. 
The rank-rank histogram (Fig.~\ref{fig:rtd_1gram_mimic} left) is useful for evaluating how the word-frequency distributions (Fig.~\ref{fig:rtd_1gram_mimic} right) are similar or disjoint among the two classes, and in the process visually inspecting the fit of the tunable parameter $\alpha$, which modulates the impact of lowly-ranked $n$-grams. 
See Figs.~\ref{fig:2gram_nogen},~\ref{fig:1grams_nogen},~\ref{fig:3grams_nogen},~and~\ref{fig:2grams_gen} for additional allotaxonographs, including 2- and 3-grams.

In the case of our gender classes in the medical data, we find the rank distributions to be more similar than disjoint and visually confirm that $\alpha=1/3$ is an acceptable setting (by examining the relation between contour lines and rank-rank distribution). 

More specifically, in our case study gendered language is highlighted by calculating the RTD values for male and female patient notes. 
We present results from applying our RTD method to $1$-grams in the unmodified \MIMIC{} dataset in Fig.~\ref{fig:rtd_1gram_mimic}. 
Unsurprisingly, gendered pronouns appear as the greatest contribution to RTD between the two corpora. 
Further, $1$-grams regarding social characteristics such as ``husband'', ``wife'', and ``daughter'' and medically relevant terms relating to sex-specific or sex-biased conditions such as ``hysterectomy'', ``scrotal'', and ``parathyroid'' are also highlighted. 

Some of these terms may be obvious to readers---suggesting the effectiveness of this approach to capture intuitive differences. 
However the identification and ranking of other $n$-grams in terms of gendered-bias requires examination of a given dataset---perhaps indicating unintuitive relationships between terms and gendered language, or potentially indicating overfitting of this approach to specific datasets. 
The application of RTD produces, in a principled fashion, a list of target terms to remove during the debiasing process---automating the selection of biased $n$-grams and tailoring results to a specific dataset.  
Using the same RTD results from above, we apply our trimming procedure---augmenting the text by iteratively removing the most biased 1-grams.
For instance, in the \MIMIC{} data the top 268 $1$-grams account for 10\% of the total RTD identified by method---and these are the first words we trim.


\subsection{Gender classification}

As an extrinsic evaluation of our bias removal process, we present performance results for classifiers predicting membership in the two classes that we obscure through the data augmentation process. 
We posit that the performance of a classifier in this case is an important metric when determining if the data augmentation was successful in removing bias signals. 
The performance of the classifier is analogous to a real-world application under an extreme case where we are trying to predict the protected class. 

We evaluate the performance of a binary gender classifier based on BERT and Clinical BERT language models. 
As a starting point, we investigate the performance of a basic nearest neighbors classifier running on document embeddings produced with off-the-shelf language models.
The classification performance of the nearest-neighbor classifier is far better than random and speaks to the embedding space's local clustering by gendered of words in these datasets, suggesting that gender may be a major component of the words embedded within this representation space.  
The tendency for BERT, and to a lesser extent Clinical BERT, to encode gender information can be seen in the tSNE visualization of these document embeddings~(Figs.~\ref{fig:tSNE_BERT_i2b2}~and~\ref{fig:tSNE_ClinicalBERT_i2b2}). 
As seen here and in other results, Clinical BERT exhibits less gender-bias according to our metrics. 
We leave a more in-depth comparison of gender-bias in BERT and Clinical BERT to future work, but it is worth noting that different embeddings appear to have different levels of gender-bias. 
Further, clinical text data may be more or less gender-biased than everyday text.

The performance of the BERT-based nearest neighbor classifier on the gender classification task is notable (Mathews correlation coefficient of 0.69) given the language models were not fine-tuned (Table~\ref{tab:mcc_baselines}).
Using Clinical BERT embeddings result in a MCC of 0.44 for the nearest neighbor classifier---with Clinical BERT generally performing slightly worse on gender classification tasks. 

As a point of comparison, we attempt a naive approach to removing gender bias through data augmentation that involves trimming a manually selected group of 20 words. 
When we run our complete BERT classifier, with fine tuning, for 1 epoch we find that the MCC drops from 0.94 to -0.06 when we trim the manually selected words.
This patterns holds up for Clinical BERT as well. 
However, if we extend the training run to 10 epochs, we find that most of the classification performance is recovered. 
This suggests that although the manually selected terms may have some of the most prominent gender signals, removing them by no means prevents the models from learning other indicators of gender.


\begin{table}[ht]
\rowcolors{2}{lightgray}{white}
    \centering
    \begin{tabular}{c|c|c|c|c}
    {} & \multicolumn{2}{c}{\textbf{BERT}} & \multicolumn{2}{c}{\textbf{Clinical BERT}} \\
    \midrule
    \textbf{Model notes} & Gendered & No-gend. & Gendered & No-gend. \\
    \hline \hline
    Nearest neighbor & 0.69  & * & 0.44 & * \\
    \hline
    1 Epoch & 0.94 & -0.06 & 0.92 & 0.00 \\
    \hline
    10 Epochs & * &  0.88 & * & 0.56  \\
    \hline
    
    \end{tabular}
    \caption{\textbf{Mathews Correlation Coefficient for gender classification task on \iibb dataset.} BERT and Clinical BERT based models were run on the manually generated ``no gender'' test dataset (common pronouns, etc. have been removed). 
    The nearest neighbor model uses off-the-shelf models to create document embeddings, while the models run for 1 and 10 Epochs were fine tuned.}
    \label{tab:mcc_baselines}
\end{table}

On the \MIMIC{} dataset we find gender classification to be generally accurate. 
With no gender trimming applied, MCC values are greater than $0.9$ for both BERT and Clinical BERT classifiers. 
This performance is quickly degraded as we employ our trimming method (Fig.~\ref{fig:mcc_summary}A). 
When we remove $1$-grams accounting for the first $10\%$ of the RTD, we find a MCC value of  approximately 0.2 for the gender classification task.
The removal of the initial $10\%$ of rank-divergence contributions has the most impact in terms of classification performance. 
Further trimming does not reduce the performance as much until $1$-grams accounting for nearly $80\%$ of the rank-turbulence divergence are removed. 
At this point, the classifier is effectively random, with a MCC of approximately 0.

The large drop in performance for gender classification is in contrast to that of most health conditions (Fig.~\ref{fig:mcc_summary}B). 
On the health condition classification task most trim values result in negligible drops in classification performance.

\subsection{Condition Classification}

\begin{figure*}[ht]
    \centering
    \includegraphics[width=\textwidth]{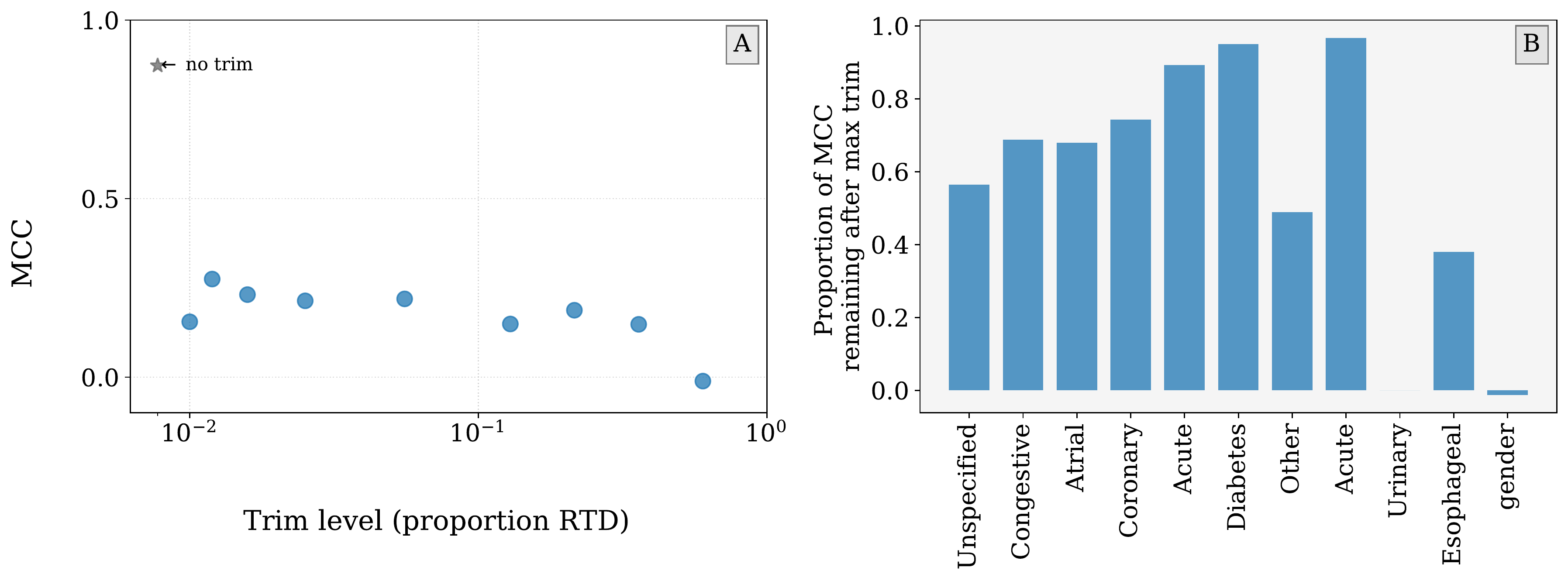}
    
    \caption{\textbf{Patient condition and gender classification performance. } 
    Using the fine-tuned Clinical BERT based model on the MIMIC{} dataset. 
    (\textbf{A}) Mathews correction coefficient scores for gender classification while varying trim levels (proportion of rank-turbulence divergence) are applied to the documents.
    (\textbf{B}) Proportion of baseline classification performance removed after maximum-trim level (70\% of total RTD) is applied to the documents.
    Of all the classification tasks, `gender' and `Urinary' experience the greatest relative decrease in classification performance. 
    However, due to the low baseline performance of Urinary ($\approx 0.2 $), the gender classification task has a notably higher absolute reduction in MCC than Urinary (or any other task). 
    See Fig.~\ref{fig:MCC_MIMIC} for full information on MCC scores for each of the health conditions. }
    \label{fig:mcc_summary}
\end{figure*}

To evaluate the impact of the bias removal process, we track the performance of classification tasks that are not explicitly linked to the two classes we are trying to protect.
Under varying levels of data augmentation we train and test multiple classification pipelines and report any degradation in performance. 
These tasks are meant to be analogous to real world applications in our domain that would require the maintenance of clinically-relevant information from the text---although we make no effort to achieve state-of-the-art results (see Table~\ref{tab:MIMIC-clinicalBERT-mcc} for baseline condition  classification performance). 

\begin{table}[ht]
\rowcolors{2}{lightgray}{white}
    \centering
    \begin{tabular}{l| l |c}
  \textbf{ICD9 Description} &  \textbf{ICD9 Code} & \textbf{MCC} \\
    \hline
   Diabetes mellitus 
   & 25000   & 0.53  \\
  Hyperlipidemia  &  2724  & 0.46 \\
  Essential hypertension  &  4019 & 0.41 \\
  Coronary atherosclerosis  &  41401 & 0.67 \\
  Atrial fibrillation  &  42731 & 0.53 \\
  Congestive heart failure  &  4280 & 0.51 \\
  Acute respiratory failure  &  51881 & 0.43 \\
  Esophageal reflux  &  53081 & 0.43 \\
  Acute kidney failure  &  5849 & 0.29 \\
  Urinary tract infection & 5990 & 0.23 
    \end{tabular}
    \caption{\textbf{Clinical BERT performance on top 10 ICD9 codes in the \MIMIC{} dataset.} }
    \label{tab:MIMIC-clinicalBERT-mcc}
\end{table}

In the specific context of our case study, we train health-condition classifiers that produce modest performance on the \MIMIC{} dataset. 
This performance is suitable for our purposes of evaluating the degradation in performance on the extrinsic task, relative to our trimming procedure. 

In the case of each health condition, we find that relative classification performance is minimally affected by the trimming procedure. 
For instance, the classifier for atrial fibrillation results in a MCC value of around 0.48 for the male patients (Fig.~\ref{fig:MCC_MIMIC}C) in the test set when no trimming is applied.  
When the minimal level of trimming is applied (10\% of RTD removed), the MCC for the males is largely unchanged, resulting in a MCC of 0.48. 
This largely holds true for most of the trimming levels, across the 10 conditions we evaluate in-depth. 
For 6 out of 10 conditions, we find that words accounting for approximately 80\% of the gender RTD need to be removed before there is a noteworthy degradation of classification performance. 
At the 80\% trim level, the gender classification task has a MCC value of approximately 0, while many other conditions maintain some predictive power.

Comparing the relative degradation in performance, we see that the proportion of MCC lost between no- and maximum-trim between 0.05 and 0.4 for most conditions (Fig.~\ref{fig:mcc_summary}B). 
The only condition with full loss of predictive power is for urinary tract infections, which one might also speculate to be related to the anatomical differences in presentation of UTIs between biological sexes. 
Although, this task also proved the most challenging and had the worst starting (no-trim) performance (MCC $\approx 0.2 $). 

\begin{figure}[h]
    \centering
    \includegraphics[width=\columnwidth]{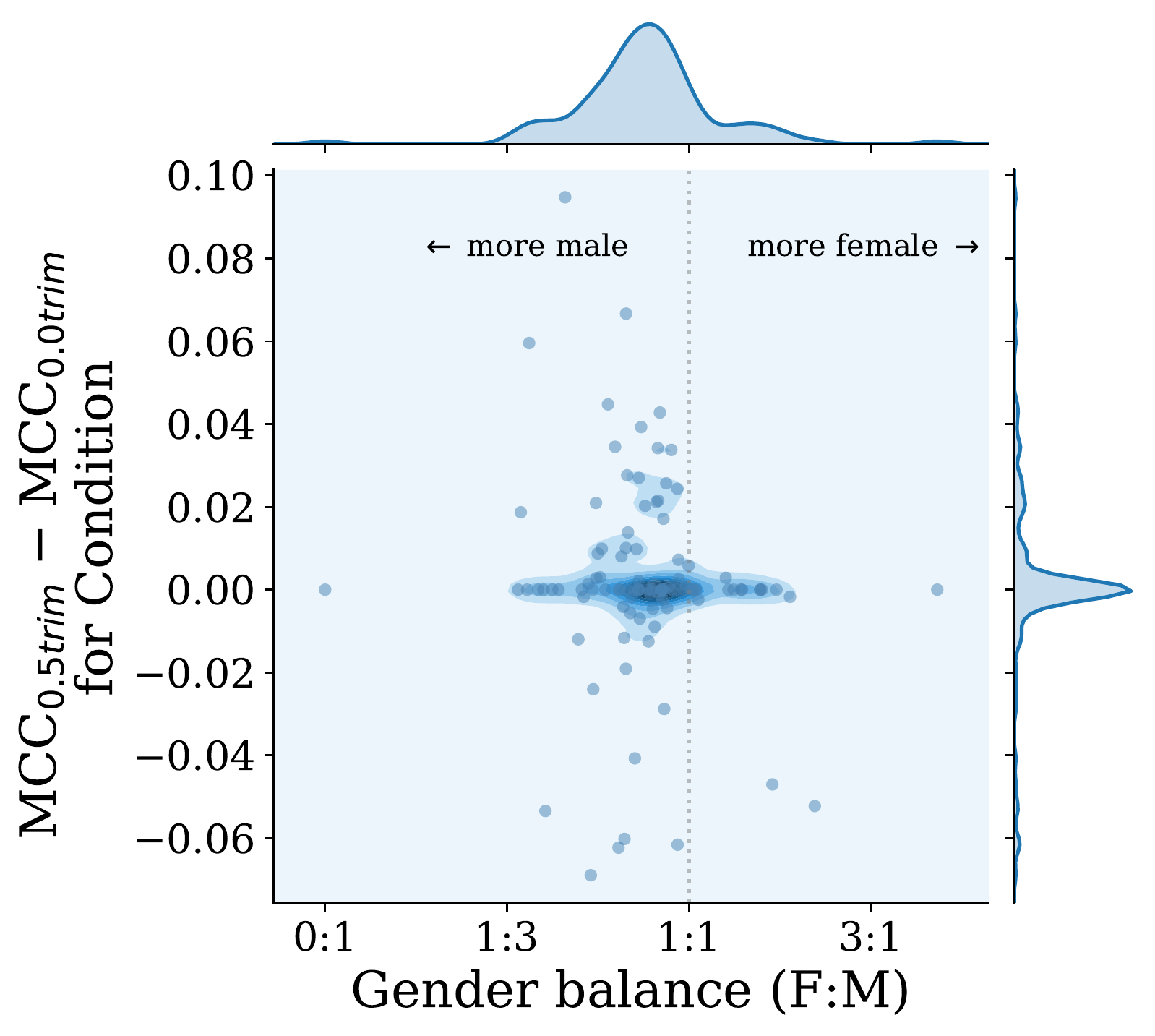}
    \caption{\textbf{Degradation in performance for Mathews correlation coefficient for condition classification of ICD9 codes with at least 1000 patients.}
    The performance degradation is presented relative to the proportion of the patients with that code who are female.
    We find little correlation between the efficacy of the condition classifier on highly augmented (trimmed) datasets and the gender balance for patients with that condition (coefficient of determination $R^2=-2.48$).   
    Values are calculated for TF-IDF based classifier and include the top 10 health conditions we evaluate elsewhere. 
    }
    \label{fig:over1000_degradation}
\end{figure}

The above results suggest that, for the conditions we examined, performance for medically relevant tasks can be preserved while reducing performance on gender classification.
There is the chance that the trimming procedure may result in biased preservation of condition classification task performance. 
To investigate this we present results from a lightweight, TF-IDF based classifier for 123 health conditions. 
We find that when we trim the top 50\% of RTD that classifiers for most conditions are relatively unaffected (Fig.~\ref{fig:over1000_degradation}).
For those conditions that do experience shifts in classification performance, any gender imbalance appears attributable related to the background gender distribution in the dataset. 

\subsection{Gender distance}

To connect the empirical data with the language models, we embed $n$-grams from our case study datasets and evaluate their bias within the word-embedding space. 
These language models have the same model-architectures that we (and many others) use when building NLP pipelines for classification and other tasks.
Bias measures based on the word-embedding space are meant to provide some indication of how debiasing techniques that are more language model-centric would operate (and what specific $n$-grams they may highlight)---keeping with our theme of interpretability while contrasting these two approaches. 

\begin{figure*}[ht!]
    \centering
    \includegraphics[width=\textwidth]{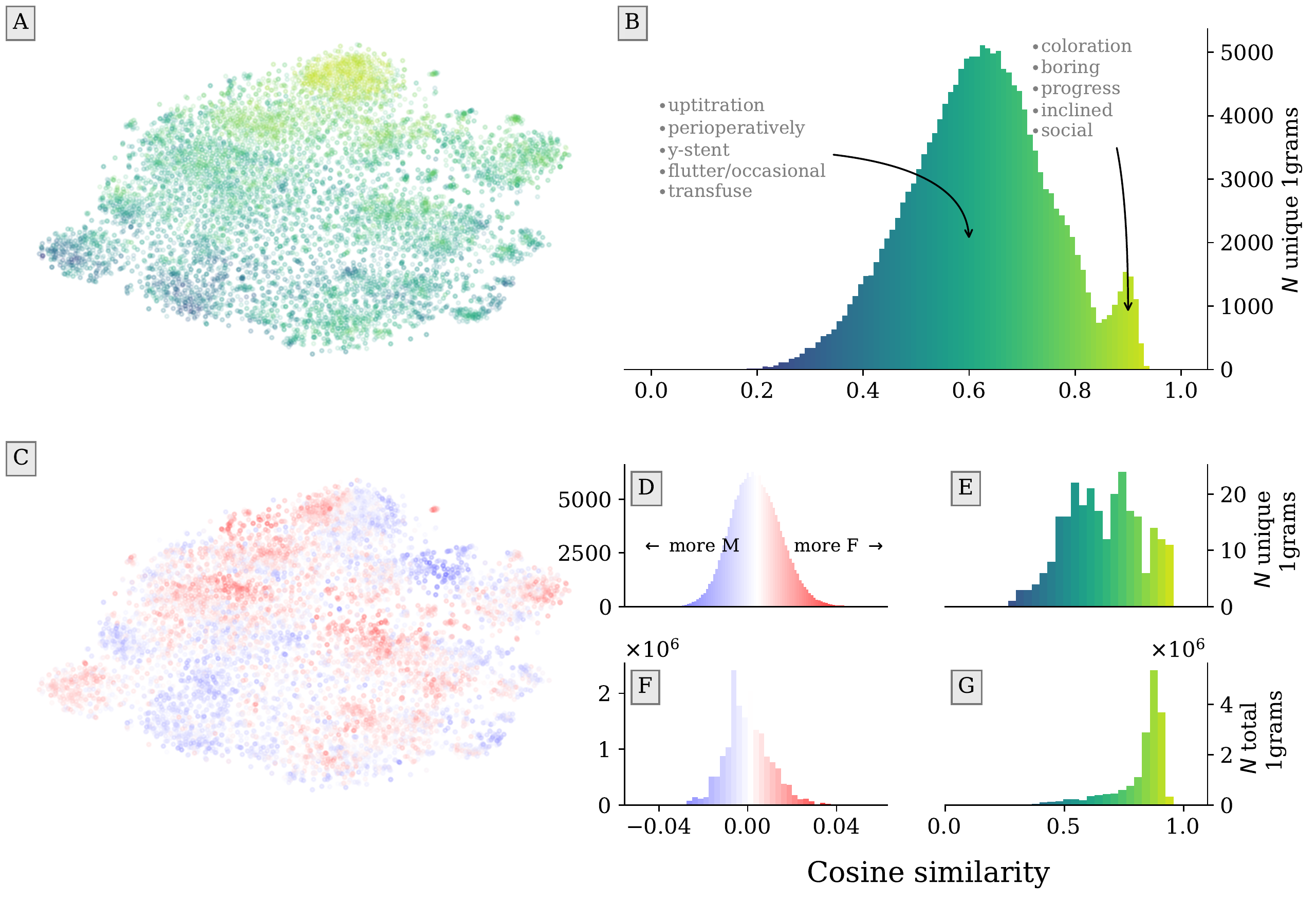}
    \caption{\textbf{Measures of gender bias in BERT word-embeddings}.
    (\textbf{A}) tSNE visualization of the BERT embedding space, colored by the maximum cosine similarity of \MIMIC{} 1grams to either male or female gendered clusters. 
    (\textbf{B}) Distribution of the maximum cosine  similarity between male or female gender clusters for 163,539 1grams appearing the \MIMIC{} corpus. Through manual inspection we find that the two clusters of cosine similarity values loosely represent more conversational English (around 0.87) and more technical language (around 0.6). 
    The words shown here were manually selected from 20 random draws for each respective region.
    (\textbf{C}) tSNE visualization of BERT embeddings space, colored by the difference in the values of cosine similarity for each word and the male and female clusters. 
    (\textbf{D}) Distribution of the differences in cosine similarity values for 1-grams and male and female clusters. 
    (\textbf{E}) Distribution maximum gendered-cluster cosine similarity scores for the 1-grams selected for removal when using the rank-turbulence divergence trim technique and targeting the top 1\% of words that contribute to overall divergence. The trimming procedure targets both common words that are considered relatively gendered by the cosine similarity measure, and less common words that are more specific to the \MIMIC{} dataset and relatively less gendered according to the cosine similarity measure. 
    (\textbf{F}) Weighted distribution of differences in cosine similarity between 1-grams and male and female clusters (same measure as (\textbf{D}), but weighted by the total number of occurrences of the 1-gram in the \MIMIC{} data). 
    (\textbf{G}) Weighted distribution of maximum cosine similarity scores between 1-grams and male or female clusters (same measure as (\textbf{B}), but weighted by the total number of occurrences of the 1-gram in the \MIMIC{} data).}
    \label{fig:cos_sim_grid}
\end{figure*}

In the context of our case study, we connect empirical data with word embeddings by presenting the distributions of cosine similarity scores for $1$-grams relative to gendered clusters in the embedding space. 
Cosine similarity scores are calculated for all 1-grams relative to clusters representing both female and male clusters (defined by $1$-grams in Table.~\ref{tab:gender_terms}). 
In our results we use both the maximum cosine similarity value relative to these clusters (i.e., the score calculated against either the female or male cluster) as well as differences in the scores for each $1$-gram relative both female and male clusters. 
Looking at the distributions of maximum cosine similarity scores for $1$-grams appearing in both the \iibb  dataset (Fig.~\ref{fig:cosine_similarity_i2b2}) and the \MIMIC{} dataset (Fig.~\ref{fig:cos_sim_grid}B) we observed a bimodal distribution of values. 
In both figures, a cluster with a mean around 0.9 is apparent as well as a cluster with a mean around 0.6.
Through manual review of the $1$-grams, we find that the cluster around 0.9 is largely comprised of more common, conversational English words whereas the cluster around 0.6 is largely comprised of medical terms. 
While there are more unique $1$-grams in the cluster of medical terms, the overall volume of word occurrences is far higher for the conversational cluster.

Referencing the cosine similarity clusters against the rank-turbulence divergence scores for the two data sets, we find that a high volume of individual $1$-grams that trimmed are present in the conversational cluster.
However, the number of unique terms there are removed for lower trim-values are spread throughout the cosine-similarity gender distribution.
For instance, when trimming the first 1\% of RTD, we find that terms are selected the more conversational cluster and more technical cluster (Fig.~\ref{fig:cos_sim_grid}E), with the former accounting for far more of the total volume of terms removed. 
The total volume of 1-grams is skewed towards the conversational cluster with terms that have higher gender similarity (Fig.~\ref{fig:cos_sim_grid}G).
The fact that the terms selected for early stages of trimming appear across the distribution of cosine similarity values illustrates the benefits of our empirical method, which is capable of selecting terms specific to a given dataset without relying on information contained in a language model.
The contrast between the RTD selection criteria and the bias present in the language model helps explain why performance on the condition classification task is minimally impacted even when a high volume of $1$-grams are removed---with RTD selecting only the most empirically biased terms.  
Using RTD-trimming there is a middle ground between obscuring gender and barely preserving performance on condition classifications---some of the more nuanced language can be retained using our method. 

\subsection{Comparison of language model and empirical bias }

Finally, we identify $n$-grams that are more biased in either the language model or in the empirical data, using RTD to divert attention away from $n$-grams that appear to exhibit similar levels of bias in both contexts. 
Put more specifically, the first application of RTD---on the empirical data and word-embeddings---ranks $n$-grams that are more male or female biased. 
The second application, the \textit{divergence-of-divergence} (RTD$^2$), ranks $n$-grams in terms of where there is most disagreement between the two bias detection approaches. 


For the \MIMIC{} dataset, we find RTD$^2$ highlights sex-specific terms, social information, and medical conditions (Table~\ref{tab:MIMIC_rtd_rtd}). 
The abbreviations of ``f'' and ``m'' for instance are rank 6288 and 244, respectively for RTD bias measures on BERT. 
Moving to RTD bias measurements in \MIMIC, ``f'' and ``m'' are the 3rd and 7th most biased terms, respectively, appearing in practically every note when describing basic demographic information for patients.
The BERT word embedding of the 1-gram ``grandmother'' has a rank of 4 but a rank of 3571 in the \MIMIC{} data---due to the fact that the 1-gram ``grandmother'' is inherently semantically gendered, but in the context of health records does not necessarily contain meaningful information on patient gender.
``Husband'' on the other hand does contain meaningful information on the patient gender (at least in the \MIMIC{} patient population), with it being rank 4 in terms of its empirical bias---the word embedding suggests it is biased, but less so with a rank of 860. 

As a final set of examples, we look at medical conditions. 
It is worth noting our choice of BERT rather than Clinical BERT most likely results in less effective word embeddings for medical terms.
``Cervical'' has a rank of 7 in the BERT bias rankings and a rank of 18374 in the empirical bias distribution---most likely owing to the split meanings in a medical context.
Conversely, ``flomax'' has a rank of 10891 for the word embedding bias, while the empirical bias rank is 11---most likely due to the gender imbalance in the incidence of conditions (e.g., kidney stones, chronic prostatitis) that flomax is often prescribed to treat.
Similarly, ``hypothyroidism'' is ranked 12 in MIMIC{} and 17831 in BERT RTD ranks, with the condition having a known increased prevalence in female-patients.
 
The high RTD$^2$ ranks for medical conditions somewhat owe to the fact that we used BERT rather than the medically-adapted Clinical BERT for these results.
For these results the choice to use the general purpose BERT rather than Clinical BERT was motivated by illustrating the discrepancies in bias rankings when using the general purpose model (with the added contrast of a shifted domain, as indicated by jargonistic medical conditions). 
When applying this type of comparison in practice, it will most likely be more beneficial to compare bias ranks with language models that are used in any final pipeline (in this case, Clinical BERT). 
Additionally, the difficulty of constructing meaningful clusters of gendered terms using technical language limits the utility of the our cosine similarity bias measure in the Clinical BERT embedding space (see Table~\ref{tab:MIMIC_clinicalBERT_rtd_rtd}).  
Inspection of the $1$-grams with high RTD$^2$ values for BERT suggests a word of caution when using general purpose word embeddings on more technical datasets, while also illustrating how specific terms that drive bias may differ between different domains.
The lesson derived by the case study of applying BERT to medical texts could be expanded to provide further caution when working in domains that do not have the benefit of fine-tuned models or where model fit may be generally poor for other reasons.






\section{Concluding Remarks}\label{sec:conclusion}
Here we present interpretable methods for detecting and reducing bias in text data. 
Using clinical notes and gender as a case study, we explore how using our methods to augment data may affect performance on classification tasks, which serve as extrinsic evaluations of the bias removal process. 
We conclude by contrasting the inherent bias present in language models with the bias we detect in our two example datasets. 
These results demonstrate that it is possible to obscure gender-features while preserving the signal needed to maintain performance on medically relevant classification tasks. 

When evaluating the differences in word use frequency in medical documents, certain intuitive results emerge: practitioners use gendered pronouns to describe patients, they note social and family status, and they encode medical conditions with known gender imbalances. 
Using our rank-turbulence divergence approach, we are able to evaluate how each of these practices, in aggregate, contribute to a divergence in word-frequency distributions between the male- and female-patient notes.
This becomes more useful as we move to identifying language that while not explicitly gendered may still be used in an unbalanced fashion in practice (for instance, non-sex specific conditions that are diagnosed more frequently in one gender). 
The results from divergence methods are useful for both understanding differences in language usage and as a debiasing technique. 

While many methods addressing debiasing language models focus on the bias present in the model itself, our empirically-based method offers stronger debiasing of the data at hand.
Modern language models are capable of detecting gender signals in a wide variety of datasets ranging from conversational to highly technical language.
Many methods for removing bias from the pre-trained language model still leave the potential of meaningful proxies in the target dataset, while also raising questions on degradation in performance. 
We believe that balancing debiasing with model performance is benefited by interpretable techniques, such as those we present here. 
For instance, our bias ranking and iterative application of divergence measures allow users to get a sense of disagreement in bias ranks for language models and empirical data.

Due to the available data we were not able to develop methods that address non-binary cases of gender bias. 
There are other methodological considerations for expanding past the binary cases~\cite{cao2019toward}, although this is an important topic for a variety of bias types~\cite{manzini2019black}.

There are further complications when moving away from tasks where associated language is not as neatly segmented. 
For instance, we show above that when evaluating language models such as BERT much of the gendered language largely appears in a readily identifiable region of the semantic space. 
As a rough heuristic: terms appearing in a medical dictionary tended to be less similar to gendered terms than terms that might appear in casual conversation.
For doctors notes, the bulk of the bias stems from words that are largely distinct from those that we expect to be most informative for medically relevant tasks. 
Further research is required to determine the efficacy of our techniques in domains where language is not as neatly semantically segmented.

Future research that applies these interpretable methods to clinical text have the opportunity to examine possible confounding factors such as patient-provider gender concordance.
Other confounding factors relating to the patient populations and broader socio-demographic factors could be addressed by replicating these trials on new data sets.
There is also the potential to research how presenting the results of our empirical bias analysis to clinicians may affect note writing practices---perhaps adding empirical examples to the growing medical school curriculum that addresses unconscious bias \cite{teal2012helping}. 

Our methods make no formal privacy guarantees nor do we claim complete removal of bias.
There is always a trade-off when seeking to balance bias reduction with overall performance, and we feel our methods will help all stakeholders make more informed decisions.
Our methodology allows stakeholders to specify the trade-off between bias reduction and performance that is best for their particular use case by selecting different trim levels and reviewing the $n$-grams removed.
Using a debiasing method that is readily interpreted by doctors, patients, and machine learning practitioners is a benefit for all involved, especially as public interest in data privacy grows.


Moving towards replacing strings rather than trimming or dropping them completely should be investigated in the future. 
More advanced data augmentation methods may be needed if we were to explore the impact of debiasing on highly tuned classification pipelines.
Holistic comparisons of string replacement techniques and other text data augmentation approaches would be worthwhile next steps. 
Further research on varying and more difficult extrinsic evaluation tasks would be helpful in evaluating how our technique generalizes.
Future work could also investigate coupling our data-driven method with methods focused on debiasing language models.

\acknowledgments 
The authors are grateful for the computing resources provided by the Vermont Advanced Computing Core 
and financial support from the Massachusetts Mutual Life Insurance Company and Google.
The views expressed are those of the authors and do not necessarily reflect the position or policy of the Department of Veterans Affairs or the United States government. 

\bibliography{references}

\clearpage

\newwrite\tempfile
\immediate\openout\tempfile=startsupp.txt
\immediate\write\tempfile{\thepage}
\immediate\closeout\tempfile

\setcounter{page}{1}
\renewcommand{\thepage}{S\arabic{page}}
\renewcommand{\thefigure}{S\arabic{figure}}
\renewcommand{\thetable}{S\arabic{table}}
\setcounter{figure}{0}
\setcounter{table}{0}

\section{Supplementary Information (SI)}

\subsection{Note selection}
\label{sec:SI_note_selection}

\begin{figure*}[tp!]
    \centering
    \includegraphics[width=\textwidth]{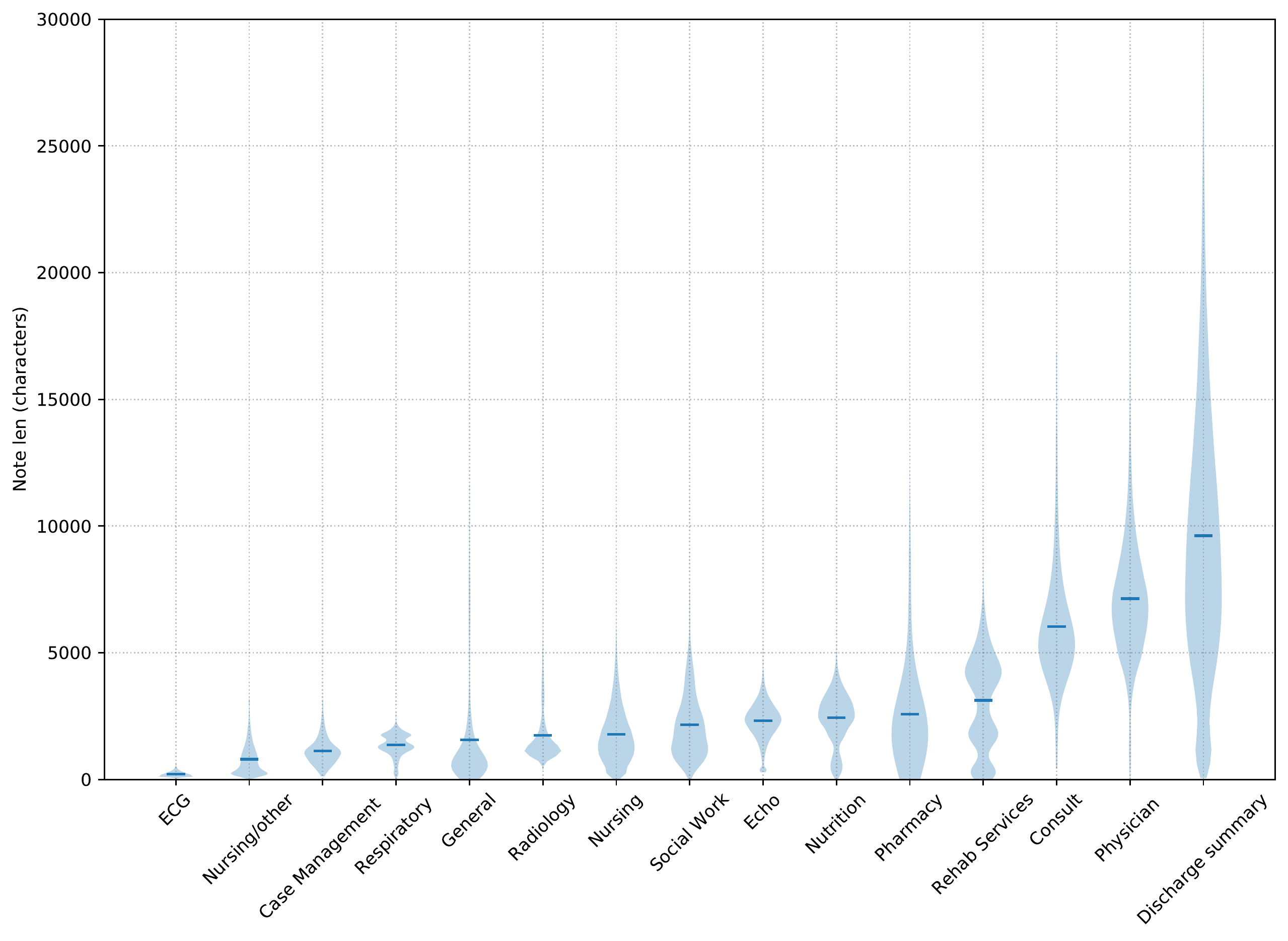}
    \caption{\textbf{Document length for MIMIC-III by note type.}}
    \label{fig:doclen_notetype}
\end{figure*}

After reviewing the note types available in the \MIMIC dataset, we determined that many types were not suitable for our task. This is due to a combination of factors including information content and note length (Fig.~\ref{fig:doclen_notetype}. 
Note types such as \texttt{radiology} often include very specific information (not indicative of broader patient health status), are shorter, and may be written in a jargonistic fashion. 
For the work outlined here we only include notes that are of the types \texttt{nursing}, \texttt{discharge summary}, and \texttt{physician}. 
In order to be included in our training and test datasets, documents must come from patients with at least three recorded documents.

\subsection{Variable length note embedding}
\label{sec:SI_variable_len_note}

When tokenized, many of notes available in the MIMIC-III dataset are longer than the 512-token maximum supported by BERT. To address this issue with experiment with truncating the note at the first 512 tokens. We also explore embedding at the sentence level (embedding with a maximum of 128 tokens) and simply  dividing the note in 512-token subsequences. In the latter two cases, we use the function outlined by Huang \textit{et al.} \cite{huang2019clinicalbert}, 

\begin{equation}
P(Y=1)  = 
\frac{P^n_{max} + P^n_{mean}n/c}
{1+n/c}
\end{equation}

where $P^n_{max}$ and $P^n_{mean}$ are the maximum and mean probabilities for the $n$ subsquences associated with a given note. Here, $c$ is a tunable parameter that is adjusted for each task. 

For our purposes, the improvement in classification performance returned by employing this technique did not merit use in our final results. 
If overall performance of our classification system were our primary objective, this may be worth further investigation.

\subsection{Hardware}

BERT and Clinical BERT models were fine-tuned on both an NVIDIA RTX 2070 (8GB VRAM) and  NVIDIA Tesla V100s (32GB VRAM).

\begin{table}[h]
    \centering
    \begin{tabular}{|l|l|}
    \hline
       \textbf{Female $1$-grams } & \textbf{Male $1$-grams} \\
       \hline
          her &  his \\
          she &   he \\
          woman &  man \\
          female & male \\
          Ms &  Mr \\
          Mrs &  him \\
          herself & himself \\
          girl & boy \\
          lady &   gentleman \\
\hline
    \end{tabular}
    \caption{\textbf{Manually selected gendered terms}.}
    \label{tab:gender_terms}
\end{table}


\begin{figure}[h!]
    \centering
    \includegraphics[width=\columnwidth]{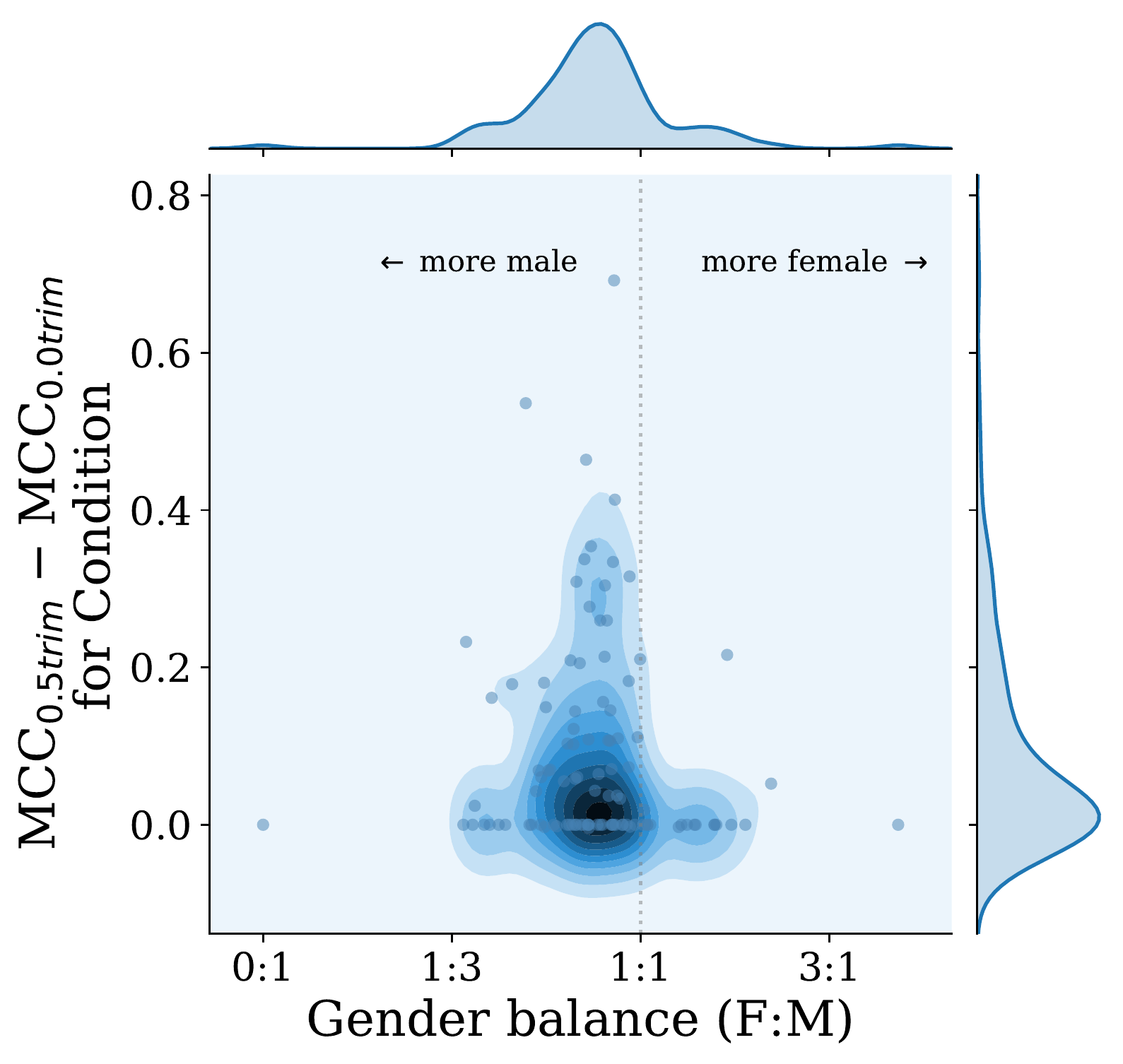}
    \caption{\textbf{Classification performance for the next 123 most frequently occurring conditions}.
    Mathews correlation coefficient for condition classification of ICD9 codes with at least 1000 patients compared to the proportion of the patients with that code who are female.
    While the most accurate classifiers tend to be fore conditions with a male bias, we observed that this is in-part due to the underlying bias in patient gender. 
    }
    \label{fig:over1000}
\end{figure}

\begin{figure*}[ht!]
    \includegraphics[width=\textwidth]{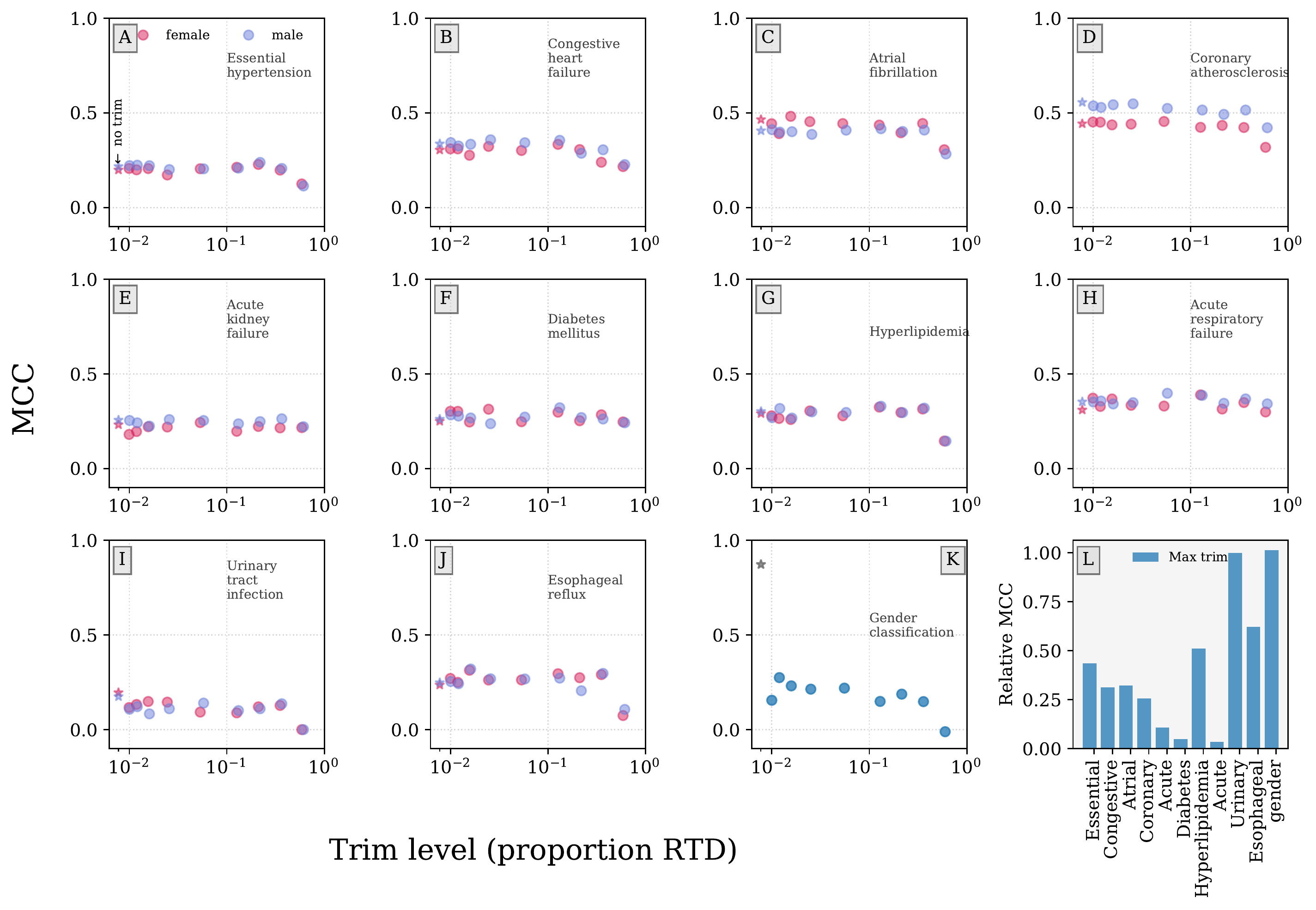} 
   \caption{ \textbf{Mathews correlation coefficient (MCC) for classification results of health conditions and patient gender with varying trim levels}. 
   Results were produced with clinicalBERT embeddings and no-token $n$-gram trimming.
   (\textbf{A}) $-$ (\textbf{J})  show MCC for the top 10 ICD9 codes present in the MIMIC data set. 
   (\textbf{K}) shows MCC for gender classification on the same population. 
   (\textbf{L}) presents a comparison of MCC results for data with no trimming and the maximum trimming level applied. 
   Values are the relative MCC, or the proportion of the best classifiers performance we lose when applying the maximum rank-turbulence divergence trimming to the data. 
   Here we see the relatively small effect of gender-based rank divergence trimming on the condition classification tasks for most conditions.
   The performance on the gender classification task is significantly degraded, even at modest trim levels, and is effectively no better than random guessing at our maximum trim level. 
   It is worth noting that many conditions are stable for most of the trimming thresholds, although we do start to see more consistent degradation of performance at the maximum trim level for a few conditions.  }
    \label{fig:MCC_MIMIC}
\end{figure*}

\begin{figure*}
    \centering
    \includegraphics[width=\textwidth]{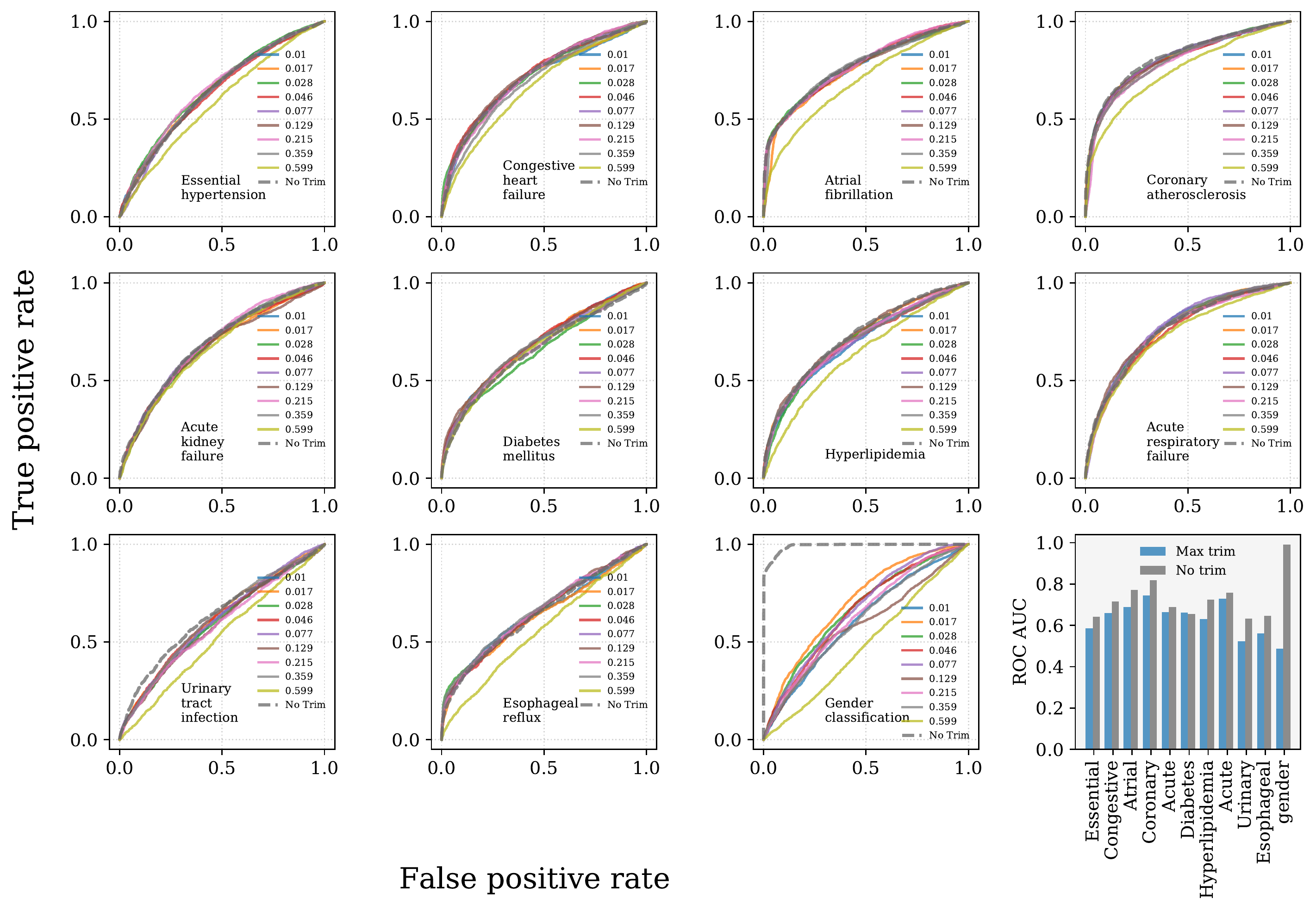}
    \caption{\textbf{ROC curves for classification task on top 10 health conditions with varying proportions of rank-turbulence divergence removed.}
    Echoing the results in Fig.~\ref{fig:MCC_MIMIC}, the gender classifier has the best performance on the `no-trim' data and experiences the greatest drop in performance when trimming in applied. 
    Under the highest trim level reported here, the gender classifier is effectively random, while few condition classifiers retain prediction capability (albeit modest).
    The bar chart show the area under the ROC curve for classifiers, by task, trained and tested with no-trimming and maximum-trimming applied.}
    \label{fig:ROC_basic}
\end{figure*}


\begin{figure*}[ht]
    \centering
    \includegraphics[width=\textwidth]{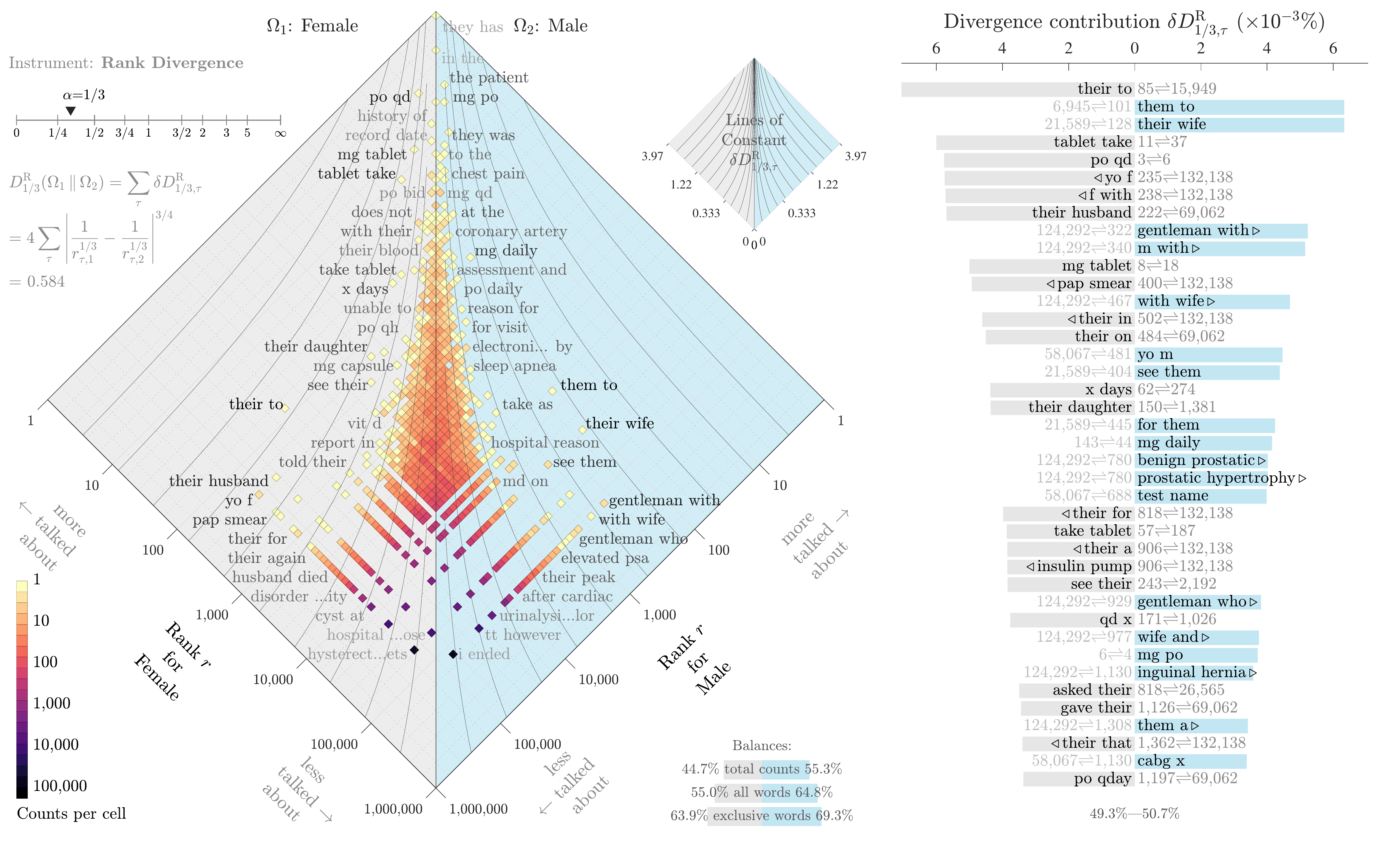}
    \caption{\textbf{Rank-turbulence divergence for 2014 i2b2 challenge.} For this figure, 2-grams have been split between genders and common gendered terms (pronouns, etc.) have been removed before calculating rank divergence. }
    \label{fig:2gram_nogen}
\end{figure*}

\begin{figure*}[h]
    \centering
    \includegraphics[width=\textwidth]{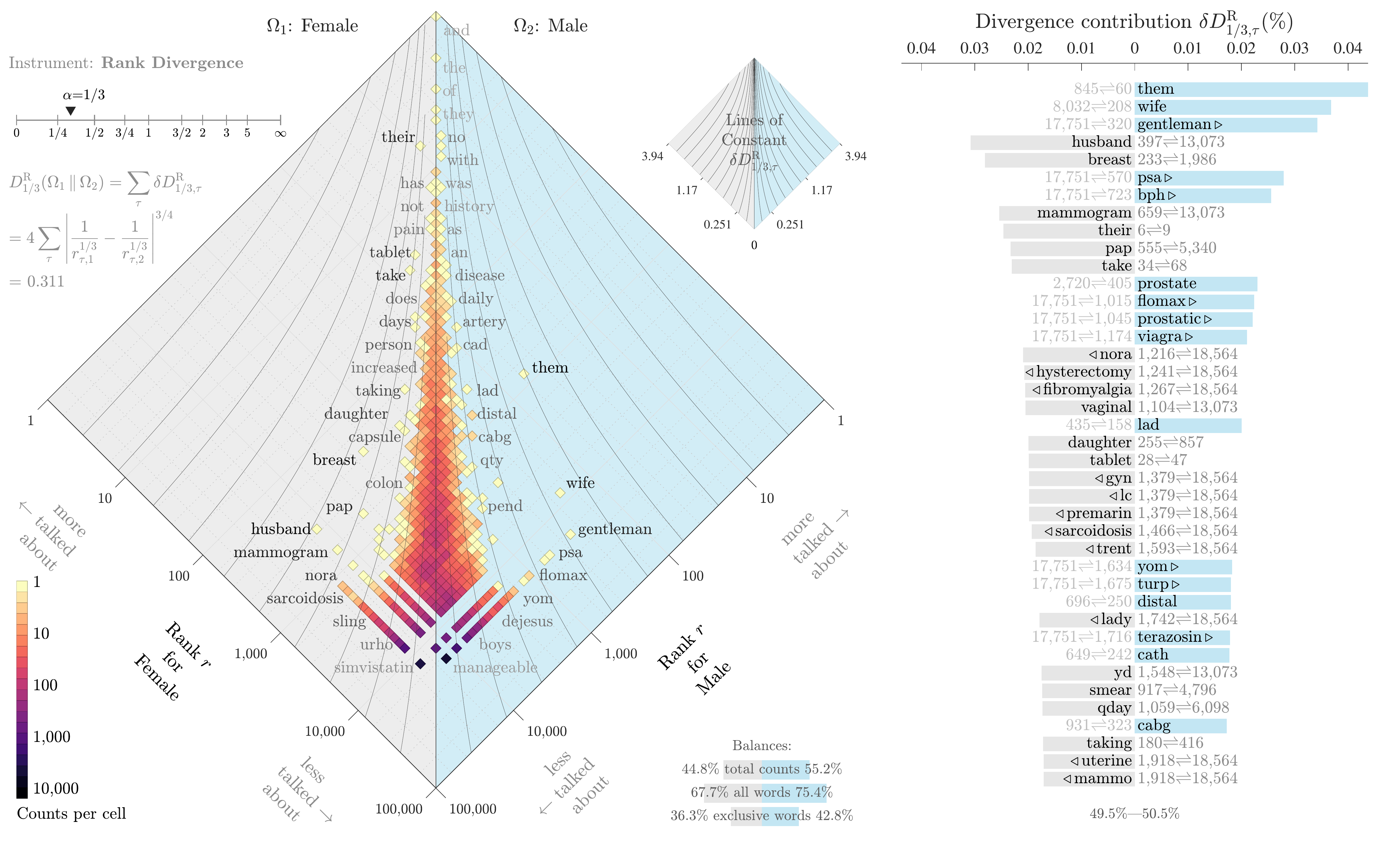}
    \caption{Rank-turbulence divergence for 2014 i2b2 challenge. For this figure, 1-grams have been split between genders and common gendered terms (pronouns, etc. see Table~\ref{tab:gender_terms}) have been removed before calculating rank divergence. }
    \label{fig:1grams_nogen}
\end{figure*}

\begin{figure*}[h]
    \centering
    \includegraphics[width=\textwidth]{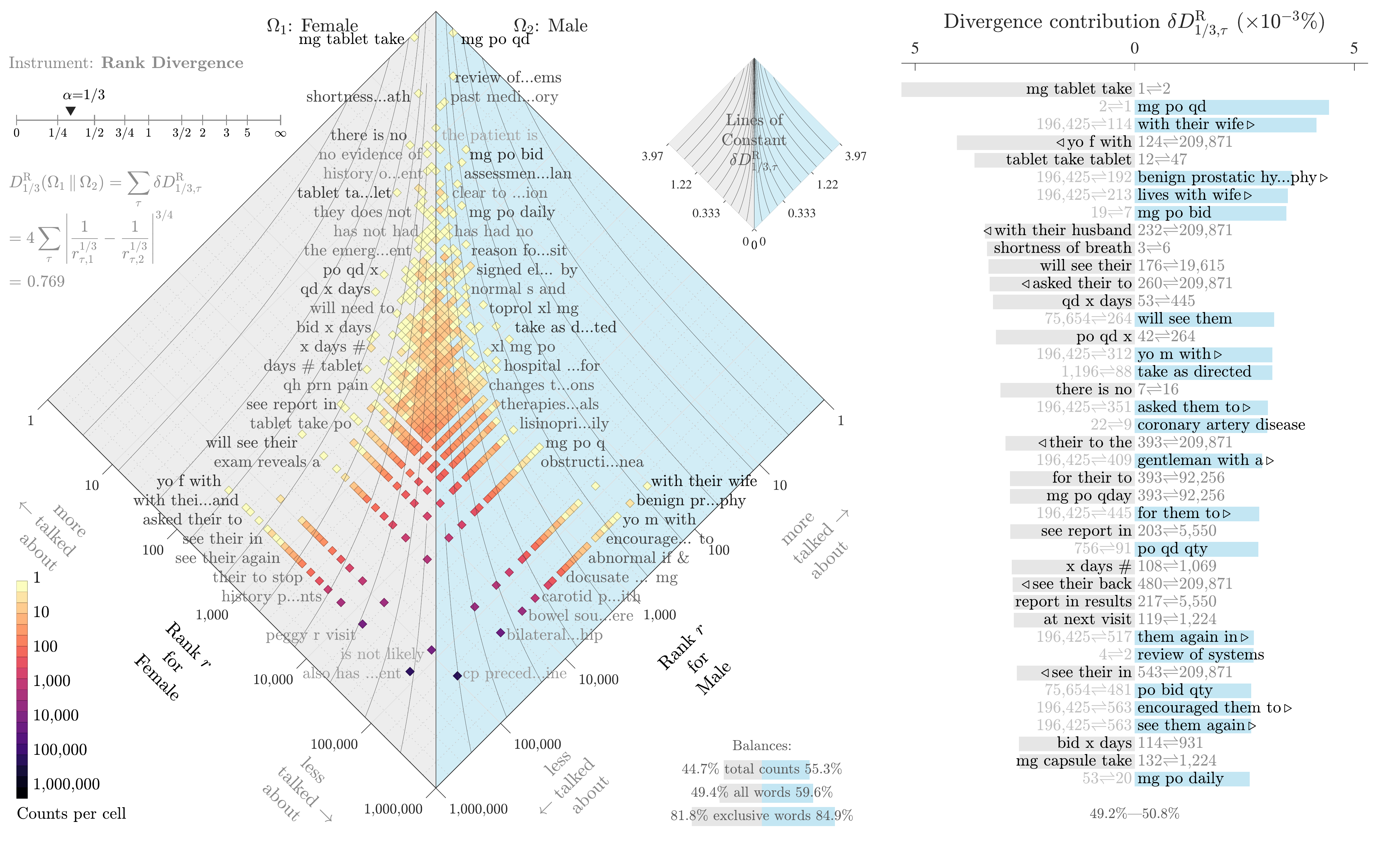}
    \caption{Rank-turbulence divergence for 2014 i2b2 challenge. For this figure, 3-grams have been split between genders and common gendered terms (pronouns, etc.) have been removed before calculating rank divergence. }
    \label{fig:3grams_nogen}
\end{figure*}

\begin{figure*}[h]
    \centering
    \includegraphics[width=\textwidth]{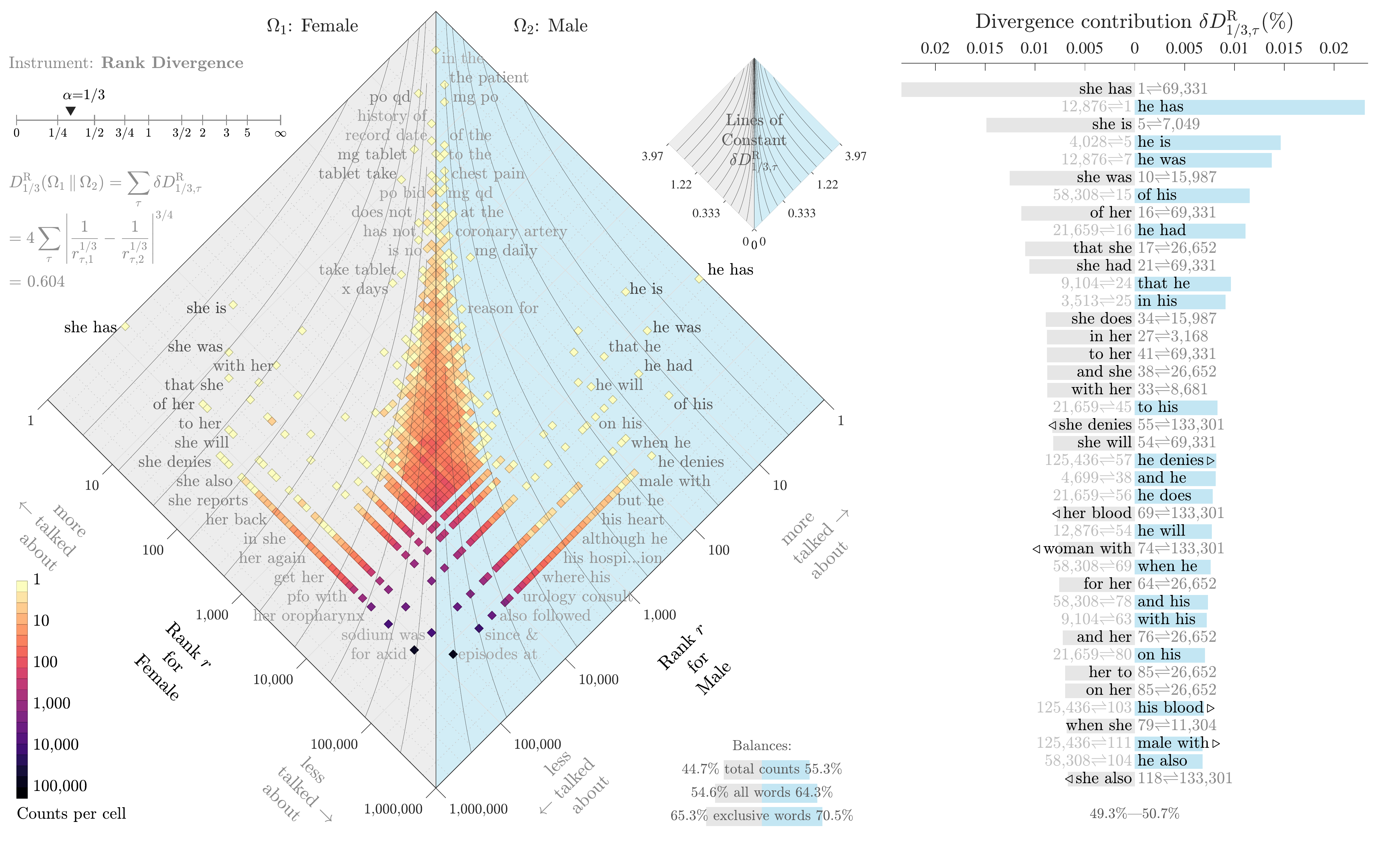}
    \caption{Rank-turbulence divergence for 2014 i2b2 challenge. For this figure, 1-grams have been split between genders. }
    \label{fig:2grams_gen}
\end{figure*}


\begin{figure}[h]
    \centering
    \includegraphics[width=\columnwidth]{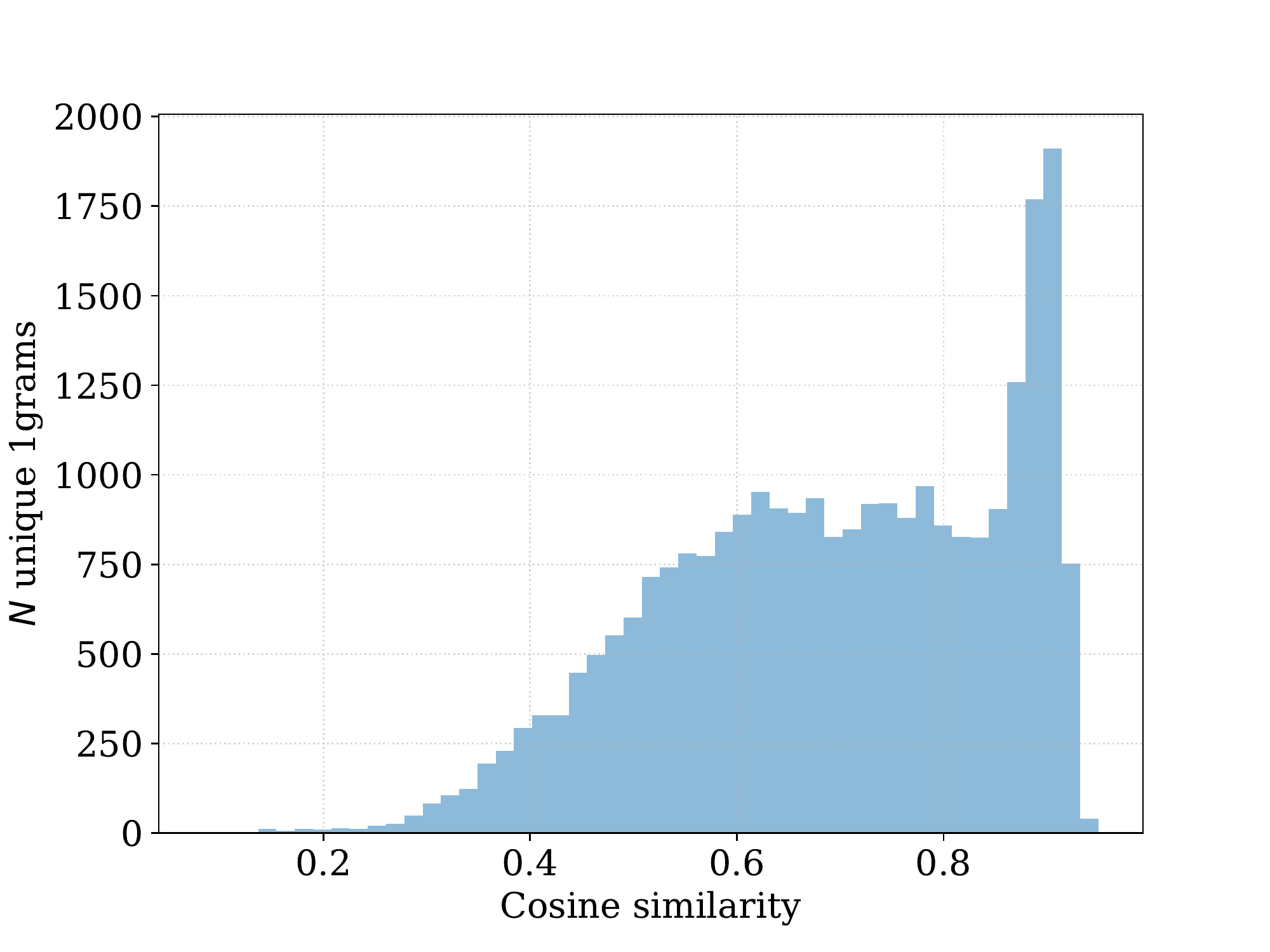}
    \caption{Maximum cosine similarity scores of BERT-base embeddings for 26,883 1grams appearing in i2b2 2014 challenge data relative to gendered clusters.
    }
    \label{fig:cosine_similarity_i2b2}
\end{figure}

\begin{figure}[h]
    \centering
    \includegraphics[width=\columnwidth]{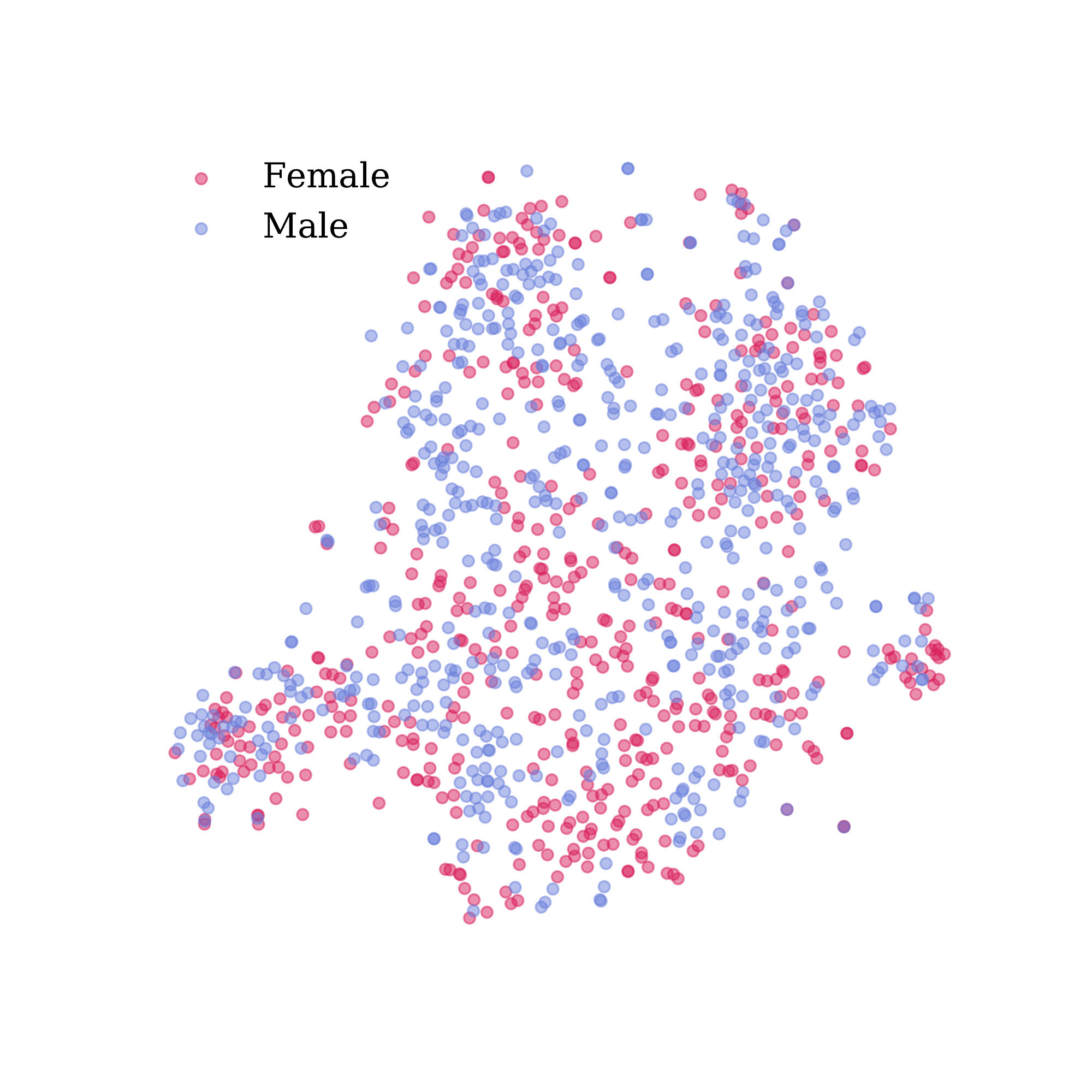}
    \caption{A tSNE embedding of i2b2 document vectors generated using a pre-trained version of Clinical BERT.}
    \label{fig:tSNE_ClinicalBERT_i2b2}
\end{figure}

\begin{figure}[h]
    \centering
    \includegraphics[width=\columnwidth]{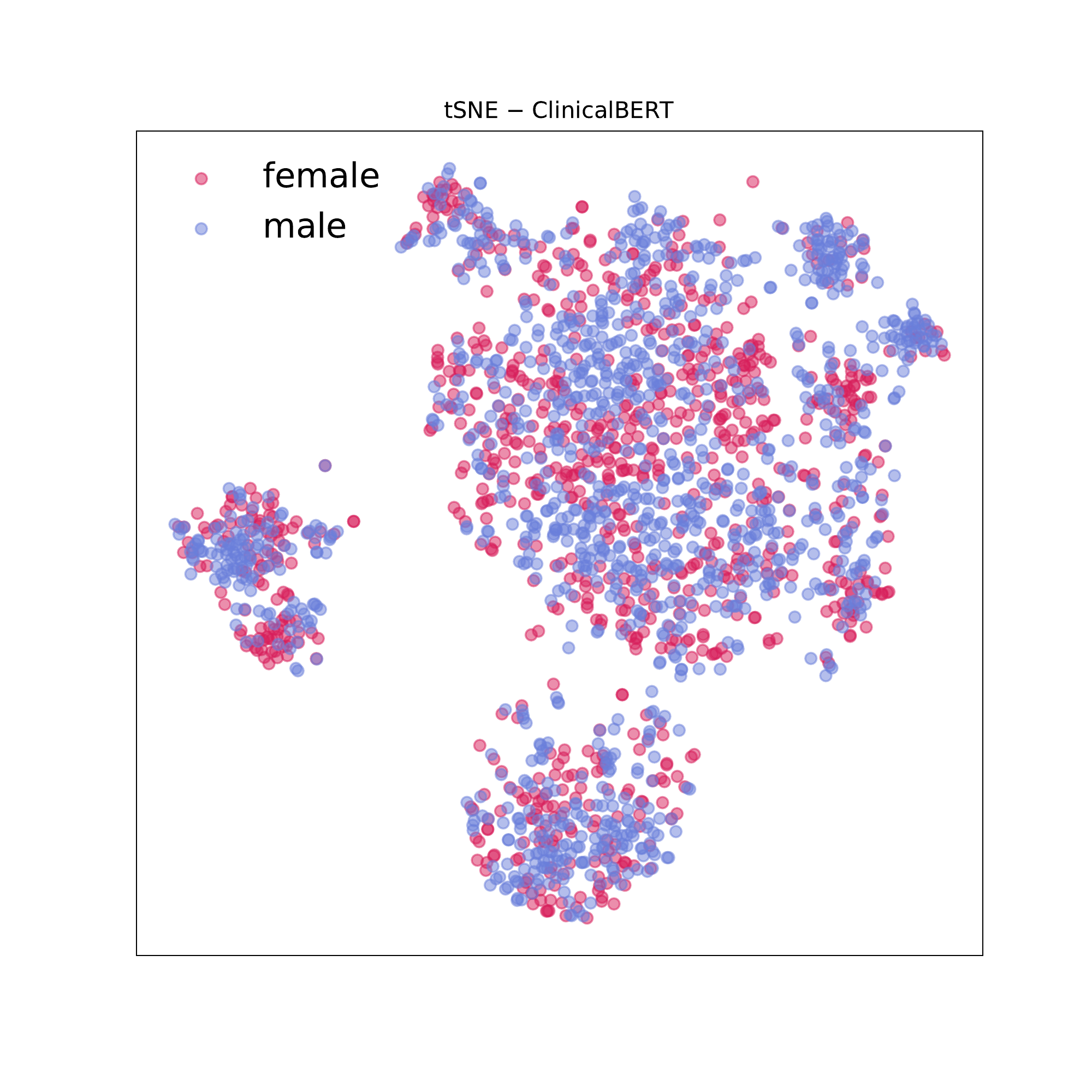}
    \caption{A tSNE embedding of MIMIC document vectors generated using a pre-trained version of Clinical BERT.}
    \label{fig:tSNE_ClinicalBERT_MIMIC_gender}
\end{figure}

\begin{figure}[hp!]
    \centering
    \includegraphics[width=\columnwidth]{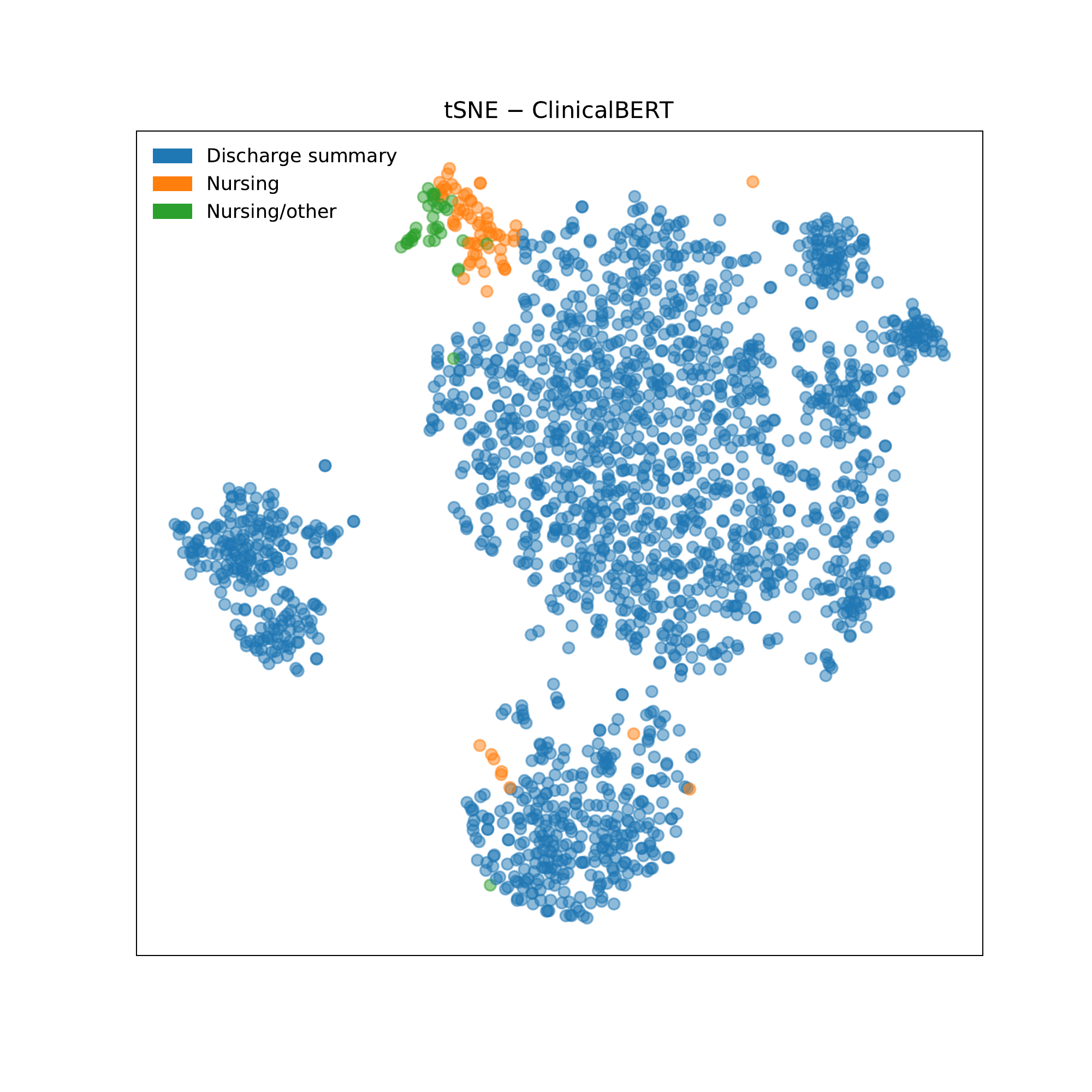}
    \caption{A tSNE embedding of MIMIC document vectors generated using a pre-trained version of Clinical BERT.}
    \label{fig:tSNE_ClinicalBERT_MIMC_notetype}
\end{figure}

\begin{figure}[h]
    \centering
    \includegraphics[width=\columnwidth]{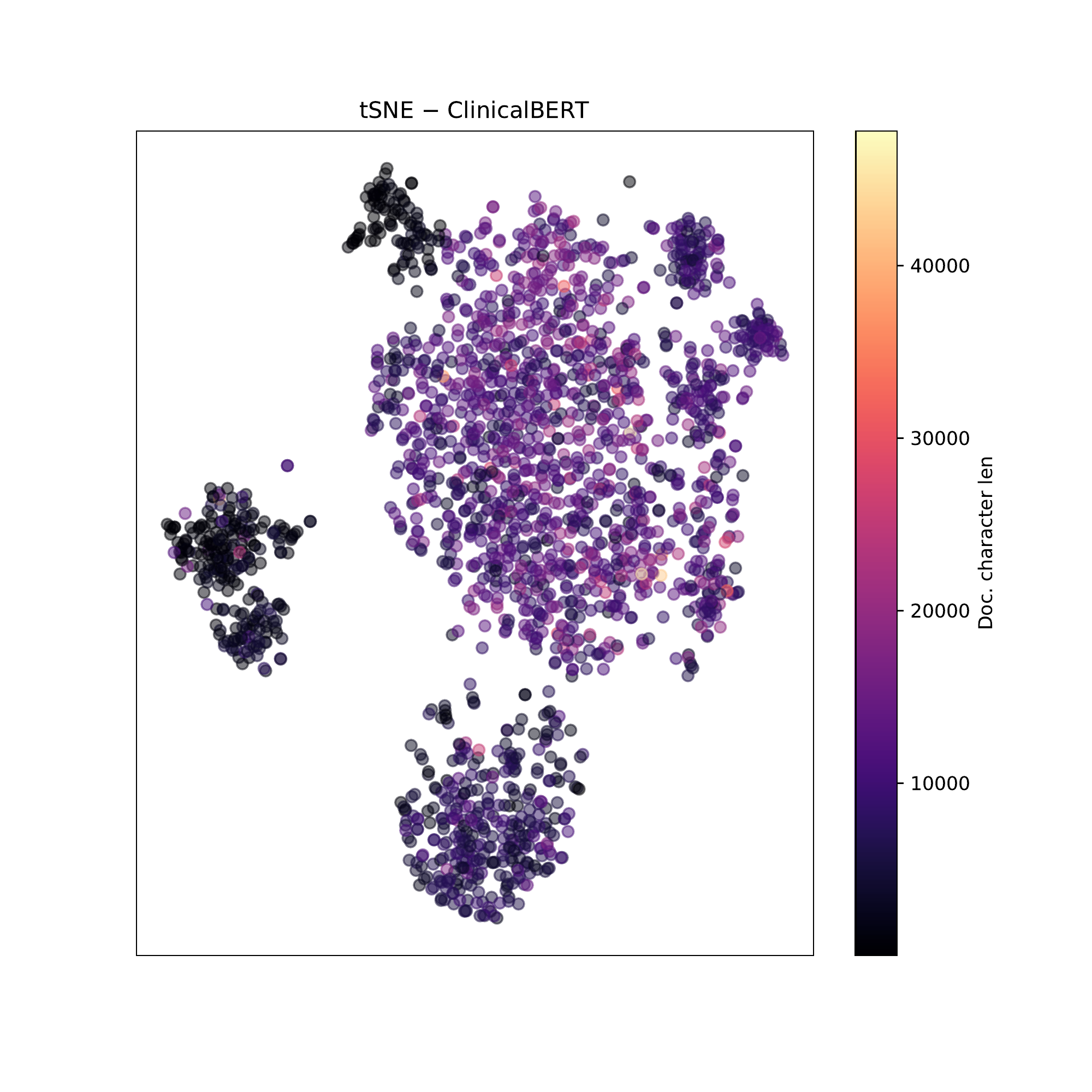}
    \caption{A tSNE embedding of MIMIC document vectors generated using a pre-trained version of Clinical BERT.}
    \label{fig:tSNE_ClinicalBERT_MIMIC_doclen}
\end{figure}



\begin{table*}[ht!]

\centering
\rowcolors{2}{lightgray}{white}
\begin{tabular}{|l|r|r|r|r|r|}

\toprule
\hline
1gram &  
\begin{tabular}{@{}c@{}}BERT \\ RTD rank\end{tabular} &  
\begin{tabular}{@{}c@{}}MIMIC \\ RTD rank\end{tabular} &  \begin{tabular}{@{}c@{}}BERT-MIMC \\ RTD rank\end{tabular} &  \begin{tabular}{@{}c@{}}MIMIC \\ F rank\end{tabular} &
\begin{tabular}{@{}c@{}}MIMIC \\ M rank\end{tabular} \\
\midrule
\hline
sexually       &            2.0 &         16755.5 &           1.0 &      10607.0 &    10373.5 \\
biggest        &            3.0 &          7594.5 &           2.0 &      17520.5 &    21172.5 \\
f              &         6288.0 &             3.0 &           3.0 &        251.0 &     1719.0 \\
infected       &            5.0 &         16076.0 &           4.0 &       3475.0 &     3402.5 \\
grandmother    &            4.0 &          3517.0 &           5.0 &       6090.0 &     7643.0 \\
cervical       &            7.0 &         18374.0 &           6.0 &       1554.0 &     1551.5 \\
m              &          244.0 &             2.0 &           7.0 &       2103.0 &      249.0 \\
sister         &            6.0 &          4153.0 &           8.0 &       1213.5 &     1365.0 \\
husband        &          860.0 &             4.0 &           9.0 &        495.0 &     5114.0 \\
teenage        &            9.0 &          5513.0 &          10.0 &      15449.0 &    19594.0 \\
trouble        &           12.0 &         10119.0 &          11.0 &       3778.5 &     4095.5 \\
brother        &           10.0 &          1921.0 &          12.0 &       1925.0 &     1598.0 \\
teenager       &            8.0 &           936.5 &          13.0 &      12928.5 &    21172.5 \\
connected      &           16.0 &         16682.0 &          14.0 &       5184.5 &     5089.5 \\
shaky          &           11.0 &          2341.0 &          15.0 &      10607.0 &     7872.0 \\
my             &           15.0 &          8198.0 &          16.0 &       2395.0 &     2624.5 \\
breast         &         3397.0 &             6.0 &          17.0 &       1075.0 &     4673.5 \\
expelled       &           19.0 &         14652.5 &          18.0 &      20814.0 &    19594.0 \\
them           &           18.0 &         11337.0 &          19.0 &       1738.5 &     1832.0 \\
prostate       &         2196.0 &             5.0 &          20.0 &       9436.0 &     1576.5 \\
immune         &           20.0 &         15043.5 &          21.0 &       9119.0 &     8753.0 \\
daughter       &            1.0 &            16.0 &          22.0 &        463.0 &      801.5 \\
initial        &           23.0 &         11433.0 &          23.0 &        632.5 &      610.0 \\
ovarian        &        16374.0 &             8.0 &          24.0 &       3137.0 &    14082.0 \\
recovering     &           24.0 &         11867.0 &          25.0 &       5351.5 &     5749.5 \\
abnormal       &           25.0 &          9010.0 &          26.0 &       1849.5 &     1994.0 \\
alcoholic      &           17.0 &          1136.0 &          27.0 &       3885.5 &     2952.0 \\
obvious        &           26.0 &         10154.0 &          28.0 &       2725.5 &     2540.5 \\
huge           &           28.0 &         13683.5 &          29.0 &       8069.0 &     7643.0 \\
dirty          &           29.0 &         13264.5 &          30.0 &       8613.5 &     9179.5 \\
suv            &           27.0 &          5911.0 &          31.0 &      14704.0 &    18377.5 \\
container      &           31.0 &         17776.0 &          32.0 &      12090.5 &    12214.5 \\
flomax         &        10891.0 &            11.0 &          33.0 &      17520.5 &     4095.5 \\
sisters        &           32.0 &         17680.5 &          34.0 &       4727.0 &     4687.5 \\
uterine        &         7260.0 &            10.0 &          35.0 &       4263.0 &    19594.0 \\
hypothyroidism &        17831.0 &            12.0 &          36.0 &       1239.5 &     2920.0 \\
dried          &           34.0 &         15661.0 &          37.0 &       5124.5 &     4982.0 \\
osteoporosis   &        18010.0 &            13.0 &          38.0 &       2354.0 &     7003.0 \\
breasts        &         4727.0 &             9.0 &          39.0 &       4394.5 &    21172.5 \\
certain        &           37.0 &         18020.5 &          40.0 &       7973.0 &     8023.0 \\
i              &           30.0 &          4601.0 &          41.0 &        403.0 &      435.5 \\
restless       &           21.0 &          1144.0 &          42.0 &       1366.0 &     1123.0 \\
wife           &           13.0 &             1.0 &          43.0 &       5545.0 &      245.5 \\
sle            &         8083.0 &            14.0 &          44.0 &       3511.0 &    12504.5 \\
granddaughter  &           14.0 &           210.0 &          45.0 &       4656.5 &     7872.0 \\
localized      &           47.0 &         14357.5 &          46.0 &       6136.0 &     6405.0 \\
ciwa           &         7921.0 &            15.0 &          47.0 &       3106.5 &     1341.0 \\
honey          &           44.0 &         11076.0 &          48.0 &      10868.5 &     9861.0 \\
coronary       &        11570.0 &            18.0 &          49.0 &        560.0 &      349.0 \\
systemic       &           41.0 &          6361.0 &          50.0 &       3696.0 &     4214.0 \\

\bottomrule
\hline
\end{tabular}
\caption{\textbf{Comparison of rank-turbulence divergences for gendered clusters in BERT embeddings and the MIMIC patient health records text.} 
BERT RTD ranks are calculated based on cosine similarity scores for word embedding and gendered clusters (i.e., the RTD of cosine similarity score ranks relative to male and female clusters). 
MIMIC RTD ranks are for $1$-grams from male and female clinical notes. 
``BERT-MIMIC RTD rank'' is the rankings for $1$-grams based on RTD between the first two columns---we also refer to this as RTD$^2$ (ranking divergence-of-divergence).  }
\label{tab:MIMIC_rtd_rtd}
\end{table*}


\begin{table*}[ht!]

\centering
\rowcolors{2}{lightgray}{white}
\begin{tabular}{|l|r|r|r|r|r|}

\toprule
\hline
1gram &  
\begin{tabular}{@{}c@{}}BERT \\ RTD rank\end{tabular} &  
\begin{tabular}{@{}c@{}}MIMIC \\ RTD rank\end{tabular} &  \begin{tabular}{@{}c@{}}BERT-MIMC \\ RTD rank\end{tabular} &  \begin{tabular}{@{}c@{}}MIMIC \\ F rank\end{tabular} &
\begin{tabular}{@{}c@{}}MIMIC \\ M rank\end{tabular} \\
\midrule
\hline
is             &            1.0 &         18545.5 &           1.0 &       24.0 &         24.0 \\
wife           &         3588.0 &             1.0 &           2.0 &      245.5 &       5545.0 \\
yells          &            2.0 &         11292.0 &           3.0 &    10563.0 &       9615.0 \\
looking        &            3.0 &         16700.0 &           4.0 &     3238.5 &       3289.5 \\
m              &         7181.0 &             2.0 &           5.0 &      249.0 &       2103.0 \\
kids           &            4.0 &         13675.0 &           6.0 &     8753.0 &       9266.5 \\
essentially    &            5.0 &         17864.0 &           7.0 &     2269.0 &       2279.5 \\
alter          &            6.0 &         15840.0 &           8.0 &    15764.0 &      16387.5 \\
f              &         2166.0 &             3.0 &           9.0 &     1719.0 &        251.0 \\
bumps          &            7.0 &          6936.0 &          10.0 &    13612.0 &      11417.0 \\
husband        &         2958.0 &             4.0 &          11.0 &     5114.0 &        495.0 \\
historian      &            8.0 &          7645.0 &          12.0 &     6895.0 &       6045.5 \\
insult         &            9.0 &          7241.0 &          13.0 &     8854.5 &      10361.0 \\
moments        &           10.0 &         16287.0 &          14.0 &    10563.0 &      10868.5 \\
our            &           11.0 &         16938.0 &          15.0 &     2803.0 &       2838.0 \\
asks           &           13.0 &         16147.0 &          16.0 &     7257.0 &       7449.5 \\
goes           &           15.0 &         17662.0 &          17.0 &     3653.0 &       3624.5 \\
someone        &           16.0 &         14030.0 &          18.0 &     6197.0 &       5920.5 \\
ever           &           17.0 &         10698.0 &          19.0 &     5114.0 &       5545.0 \\
prostate       &         6070.0 &             5.0 &          20.0 &     1576.5 &       9436.0 \\
breast         &         6010.0 &             6.0 &          21.0 &     4673.5 &       1075.0 \\
experiences    &           20.0 &         17062.0 &          22.0 &     9452.5 &       9615.0 \\
suffer         &           12.0 &          1463.5 &          23.0 &    10968.5 &      16387.5 \\
recordings     &           21.0 &         13249.5 &          24.0 &    12504.5 &      13440.5 \\
wore           &           23.0 &         14774.5 &          25.0 &     8854.5 &       9266.5 \\
largely        &           26.0 &         16941.0 &          26.0 &     4581.5 &       4515.5 \\
hi             &           22.0 &          6459.0 &          27.0 &     3992.5 &       4558.0 \\
et             &           27.0 &         17615.0 &          28.0 &     2079.0 &       2093.5 \\
staying        &           25.0 &         10254.5 &          29.0 &     4366.5 &       4746.5 \\
pursuing       &           18.0 &          1561.5 &          30.0 &    13612.0 &      20814.0 \\
pet            &           24.0 &          4873.0 &          31.0 &     5879.0 &       7076.5 \\
town           &           28.0 &         10038.0 &          32.0 &    10754.0 &       9615.0 \\
tire           &           30.0 &         12053.5 &          33.0 &    12834.5 &      11736.5 \\
ovarian        &        13390.0 &             8.0 &          34.0 &    14082.0 &       3137.0 \\
beef           &           19.0 &          1355.5 &          35.0 &    23307.0 &      14704.0 \\
dipping        &           34.0 &         17764.0 &          36.0 &     6371.0 &       6318.5 \\
dip            &           35.0 &         18328.0 &          37.0 &     5585.0 &       5600.0 \\
hat            &           33.0 &         10106.5 &          38.0 &    14082.0 &      12478.5 \\
flomax         &        14782.0 &            11.0 &          39.0 &     4095.5 &      17520.5 \\
punch          &           31.0 &          5064.5 &          40.0 &    11203.5 &      14033.0 \\
ease           &           38.0 &         17265.5 &          41.0 &     6471.5 &       6556.0 \\
hasn           &           36.0 &         11324.0 &          42.0 &    10373.5 &      11417.0 \\
lasts          &           39.0 &         14072.5 &          43.0 &    11203.5 &      10607.0 \\
grabbing       &           32.0 &          4294.0 &          44.0 &    10968.5 &      14033.0 \\
hypothyroidism &        17820.0 &            12.0 &          45.0 &     2920.0 &       1239.5 \\
whatever       &           37.0 &          8524.5 &          46.0 &    13200.5 &      15449.0 \\
osteoporosis   &        18276.0 &            13.0 &          47.0 &     7003.0 &       2354.0 \\
uterine        &         6519.0 &            10.0 &          48.0 &    19594.0 &       4263.0 \\
sle            &        16374.0 &            14.0 &          49.0 &    12504.5 &       3511.0 \\
dump           &           47.0 &         15439.0 &          50.0 &    10968.5 &      11417.0 \\
\bottomrule
\hline
\end{tabular}
\caption{\textbf{Comparison of rank-turbulence divergences for gendered clusters in Clinical BERT embeddings and the MIMIC patient health records text.} 
Clinical BERT RTD ranks are calculated based on cosine similarity scores for word embedding and gendered clusters (i.e., the RTD of cosine similarity score ranks relative to male and female clusters). 
MIMIC RTD ranks are for $1$-grams from male and female clinical notes. 
``BERT-MIMIC RTD rank'' is the rankings for $1$-grams based on RTD between the first two columns---we also refer to this as RTD$^2$ (ranking divergence-of-divergence).  
The presence of largely conversational terms rather than more technical, medical language owes to our defining of gender clusters through manually selected terms.}
\label{tab:MIMIC_clinicalBERT_rtd_rtd}
\end{table*}

\begin{figure}[tp!]
    \centering
    \includegraphics[width=\columnwidth]{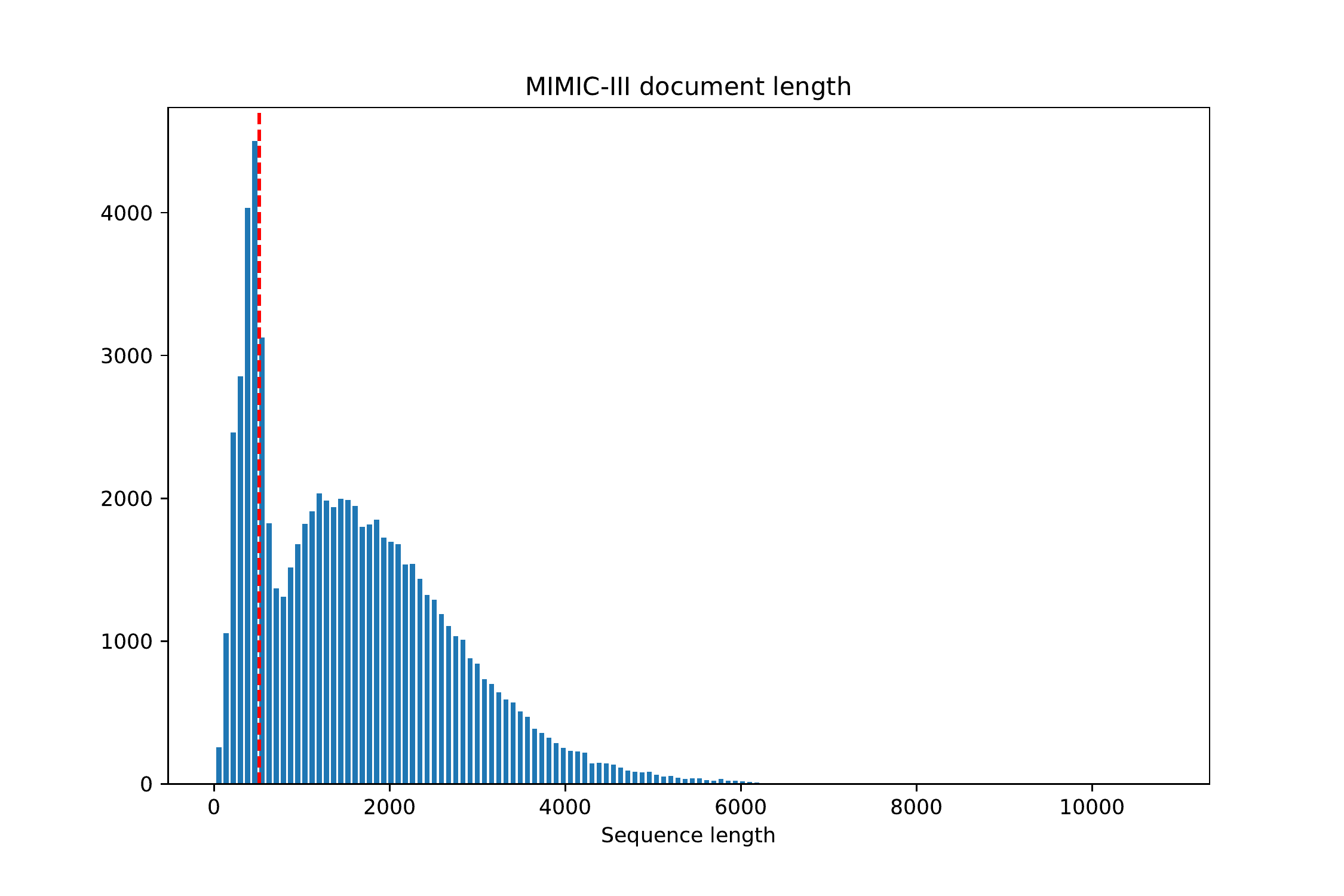}
    \caption{\textbf{Document length for \MIMIC text notes.}}
    \label{fig:doclen_mimic}
\end{figure}

\newpage


\clearpage{
\rowcolors{2}{lightgray}{white}
\begin{longtable*}{|l|r|r|r|r|}

\toprule

{} & \multicolumn{2}{c}{\textbf{Sex Count}} & \multicolumn{2}{c}{\textbf{Sex Prop.}} \\
 \textbf{ICD Description.}  &     \textbf{F} & \textbf{M} & \textbf{F} & \textbf{M}\\ 
\hline
\endhead
Personal history of malignant neoplasm of prostate &     0 &   1207 &  0.00 &  1.00 \\
Hypertrophy (benign) of prostate without urinar... &     0 &   1490 &  0.00 &  1.00 \\
Routine or ritual circumcision                     &     0 &   2016 &  0.00 &  1.00 \\
Gout, unspecified                                  &   552 &   1530 &  0.27 &  0.73 \\
Alcoholic cirrhosis of liver                       &   323 &    879 &  0.27 &  0.73 \\
Retention of urine, unspecified                    &   283 &    737 &  0.28 &  0.72 \\
Intermediate coronary syndrome                     &   466 &   1197 &  0.28 &  0.72 \\
Chronic systolic heart failure                     &   321 &    776 &  0.29 &  0.71 \\
Aortocoronary bypass status                        &   896 &   2160 &  0.29 &  0.71 \\
Other and unspecified angina pectoris              &   330 &    770 &  0.30 &  0.70 \\
Paroxysmal ventricular tachycardia                 &   548 &   1263 &  0.30 &  0.70 \\
Chronic hepatitis C without mention of hepatic ... &   380 &    838 &  0.31 &  0.69 \\
Coronary atherosclerosis of unspecified type of... &   479 &   1015 &  0.32 &  0.68 \\
Percutaneous transluminal coronary angioplasty ... &   889 &   1836 &  0.33 &  0.67 \\
Portal hypertension                                &   332 &    675 &  0.33 &  0.67 \\
Surgical operation with anastomosis, bypass, or... &   406 &    805 &  0.34 &  0.66 \\
Coronary atherosclerosis of native coronary artery &  4322 &   8107 &  0.35 &  0.65 \\
Old myocardial infarction                          &  1156 &   2122 &  0.35 &  0.65 \\
Acute on chronic systolic heart failure            &   406 &    737 &  0.36 &  0.64 \\
Cardiac complications, not elsewhere classified    &   847 &   1496 &  0.36 &  0.64 \\
Atrial flutter                                     &   444 &    773 &  0.36 &  0.64 \\
Paralytic ileus                                    &   394 &    678 &  0.37 &  0.63 \\
Chronic kidney disease, unspecified                &  1265 &   2170 &  0.37 &  0.63 \\
Personal history of tobacco use                    &  1042 &   1769 &  0.37 &  0.63 \\
Pneumonitis due to inhalation of food or vomitus   &  1369 &   2311 &  0.37 &  0.63 \\
Tobacco use disorder                               &  1251 &   2107 &  0.37 &  0.63 \\
Obstructive sleep apnea (adult)(pediatric)         &   891 &   1489 &  0.37 &  0.63 \\
Cirrhosis of liver without mention of alcohol      &   486 &    801 &  0.38 &  0.62 \\
Hypertensive chronic kidney disease, unspecifie... &  1300 &   2121 &  0.38 &  0.62 \\
Diabetes with neurological manifestations, type... &   438 &    700 &  0.38 &  0.62 \\
Other primary cardiomyopathies                     &   664 &   1045 &  0.39 &  0.61 \\
Cardiac arrest                                     &   542 &    819 &  0.40 &  0.60 \\
Peripheral vascular disease, unspecified           &   564 &    837 &  0.40 &  0.60 \\
Hyperpotassemia                                    &   874 &   1295 &  0.40 &  0.60 \\
Bacteremia                                         &   599 &    879 &  0.41 &  0.59 \\
Other and unspecified hyperlipidemia               &  3537 &   5153 &  0.41 &  0.59 \\
Thrombocytopenia, unspecified                      &  1255 &   1810 &  0.41 &  0.59 \\
Pure hypercholesterolemia                          &  2436 &   3494 &  0.41 &  0.59 \\
Pressure ulcer, lower back                         &   530 &    759 &  0.41 &  0.59 \\
Subendocardial infarction, initial episode of care &  1262 &   1793 &  0.41 &  0.59 \\
Acute kidney failure with lesion of tubular nec... &   945 &   1342 &  0.41 &  0.59 \\
Acute and subacute necrosis of liver               &   441 &    626 &  0.41 &  0.59 \\
Hypertensive chronic kidney disease, unspecifie... &  1091 &   1539 &  0.41 &  0.59 \\
Hemorrhage complicating a procedure                &   637 &    898 &  0.41 &  0.59 \\
Cardiogenic shock                                  &   480 &    674 &  0.42 &  0.58 \\
Aortic valve disorders                             &  1069 &   1481 &  0.42 &  0.58 \\
Polyneuropathy in diabetes                         &   667 &    917 &  0.42 &  0.58 \\
Other postoperative infection                      &   503 &    683 &  0.42 &  0.58 \\
Respiratory distress syndrome in newborn           &   559 &    755 &  0.43 &  0.57 \\
Cardiac pacemaker in situ                          &   592 &    798 &  0.43 &  0.57 \\
Atrial fibrillation                                &  5512 &   7379 &  0.43 &  0.57 \\
Pulmonary collapse                                 &   931 &   1234 &  0.43 &  0.57 \\
Delirium due to conditions classified elsewhere    &   622 &    823 &  0.43 &  0.57 \\
Diabetes mellitus without mention of complicati... &  3902 &   5156 &  0.43 &  0.57 \\
Hemorrhage of gastrointestinal tract, unspecified  &   602 &    795 &  0.43 &  0.57 \\
Other and unspecified coagulation defects          &   438 &    578 &  0.43 &  0.57 \\
Acute kidney failure, unspecified                  &  3941 &   5178 &  0.43 &  0.57 \\
End stage renal disease                            &   836 &   1090 &  0.43 &  0.57 \\
\newpage
Accidents occurring in residential institution     &   456 &    583 &  0.44 &  0.56 \\
Single liveborn, born in hospital, delivered by... &  1220 &   1538 &  0.44 &  0.56 \\
Sepsis                                             &   563 &    709 &  0.44 &  0.56 \\
Hyperosmolality and/or hypernatremia               &  1009 &   1263 &  0.44 &  0.56 \\
Other specified surgical operations and procedu... &   600 &    750 &  0.44 &  0.56 \\
Severe sepsis                                      &  1746 &   2166 &  0.45 &  0.55 \\
Unspecified protein-calorie malnutrition           &   562 &    697 &  0.45 &  0.55 \\
Long-term (current) use of insulin                 &  1138 &   1400 &  0.45 &  0.55 \\
Long-term (current) use of anticoagulants          &  1709 &   2097 &  0.45 &  0.55 \\
Other iatrogenic hypotension                       &   953 &   1168 &  0.45 &  0.55 \\
Anemia in chronic kidney disease                   &   623 &    761 &  0.45 &  0.55 \\
Intracerebral hemorrhage                           &   618 &    749 &  0.45 &  0.55 \\
Unspecified essential hypertension                 &  9370 &  11333 &  0.45 &  0.55 \\
Acute posthemorrhagic anemia                       &  2072 &   2480 &  0.46 &  0.54 \\
Unspecified septicemia                             &  1702 &   2023 &  0.46 &  0.54 \\
Chronic airway obstruction, not elsewhere class... &  2027 &   2404 &  0.46 &  0.54 \\
Pneumonia, organism unspecified                    &  2223 &   2616 &  0.46 &  0.54 \\
Septic shock                                       &  1189 &   1397 &  0.46 &  0.54 \\
Other convulsions                                  &   892 &   1042 &  0.46 &  0.54 \\
Other specified procedures as the cause of abno... &   693 &    809 &  0.46 &  0.54 \\
Diarrhea                                           &   484 &    565 &  0.46 &  0.54 \\
Hematoma complicating a procedure                  &   566 &    658 &  0.46 &  0.54 \\
Acute respiratory failure                          &  3473 &   4024 &  0.46 &  0.54 \\
Other specified cardiac dysrhythmias               &  1137 &   1316 &  0.46 &  0.54 \\
Need for prophylactic vaccination and inoculati... &  2680 &   3099 &  0.46 &  0.54 \\
Personal history of transient ischemic attack (... &   498 &    574 &  0.46 &  0.54 \\
Neonatal jaundice associated with preterm delivery &  1052 &   1212 &  0.46 &  0.54 \\
Observation for suspected infectious condition     &  2570 &   2949 &  0.47 &  0.53 \\
Congestive heart failure, unspecified              &  6106 &   7005 &  0.47 &  0.53 \\
Hypovolemia                        hyponatremia    &   641 &    733 &  0.47 &  0.53 \\
Single liveborn, born in hospital, delivered wi... &  1668 &   1898 &  0.47 &  0.53 \\
Unspecified pleural effusion                       &  1281 &   1453 &  0.47 &  0.53 \\
Acidosis                                           &  2127 &   2401 &  0.47 &  0.53 \\
Esophageal reflux                                  &  2990 &   3336 &  0.47 &  0.53 \\
Encounter for palliative care                      &   485 &    535 &  0.48 &  0.52 \\
Hyposmolality and/or hyponatremia                  &  1445 &   1594 &  0.48 &  0.52 \\
Iron deficiency anemia secondary to blood loss ... &   482 &    530 &  0.48 &  0.52 \\
Hypoxemia                                          &   625 &    673 &  0.48 &  0.52 \\
Mitral valve disorders                             &  1416 &   1510 &  0.48 &  0.52 \\
Primary apnea of newborn                           &   506 &    537 &  0.49 &  0.51 \\
Hypotension, unspecified                           &   996 &   1055 &  0.49 &  0.51 \\
Personal history of venous thrombosis and embolism &   786 &    826 &  0.49 &  0.51 \\
Obesity, unspecified                               &   744 &    767 &  0.49 &  0.51 \\
Intestinal infection due to Clostridium difficile  &   716 &    728 &  0.50 &  0.50 \\
Obstructive chronic bronchitis with (acute) exa... &   598 &    600 &  0.50 &  0.50 \\
Anemia of other chronic disease                    &   550 &    543 &  0.50 &  0.50 \\
Anemia, unspecified                                &  2729 &   2677 &  0.50 &  0.50 \\
Dehydration                                        &   704 &    681 &  0.51 &  0.49 \\
Other chronic pulmonary heart diseases             &  1101 &   1047 &  0.51 &  0.49 \\
Do not resuscitate status                          &   694 &    633 &  0.52 &  0.48 \\
Depressive disorder, not elsewhere classified      &  1888 &   1543 &  0.55 &  0.45 \\
Morbid obesity                                     &   648 &    522 &  0.55 &  0.45 \\
Iron deficiency anemia, unspecified                &   657 &    514 &  0.56 &  0.44 \\
Chronic diastolic heart failure                    &   708 &    532 &  0.57 &  0.43 \\
Hypopotassemia                                     &   816 &    609 &  0.57 &  0.43 \\
Anxiety state, unspecified                         &   944 &    636 &  0.60 &  0.40 \\
Dysthymic disorder                                 &   663 &    446 &  0.60 &  0.40 \\
Asthma, unspecified type, unspecified              &  1317 &    878 &  0.60 &  0.40 \\
Urinary tract infection, site not specified        &  4027 &   2528 &  0.61 &  0.39 \\
Other persistent mental disorders due to condit... &   698 &    428 &  0.62 &  0.38 \\
Acute on chronic diastolic heart failure           &   779 &    441 &  0.64 &  0.36 \\
Unspecified acquired hypothyroidism                &  3307 &   1610 &  0.67 &  0.33 \\
Osteoporosis, unspecified                          &  1637 &    310 &  0.84 &  0.16 \\
Personal history of malignant neoplasm of breast   &  1259 &     18 &  0.99 &  0.01 \\
\bottomrule 
\caption{ \textbf{Condition name and gender balance for the ICD9 codes with at least 1000 observations in the \MIMIC{} dataset.}}
\label{tab:top1000MIMIC}
\end{longtable*}
}

\end{document}